\documentclass[12pt]{article} 
\usepackage{styleFile}

\title{Scalable extensions to given-data Sobol' index estimators}
\author[1]{Teresa Portone\thanks{Corresponding author: tporton@sandia.gov}}
\author[1]{Bert Debusschere}
\author[1]{Samantha Yang}
\author[1]{Emiliano Islas-Quinones}
\author[1]{T. Patrick Xiao}

\affil[1]{Sandia National Laboratories}
\date{}

\begin{document}
\maketitle

\begin{abstract}
Given-data methods for variance-based sensitivity analysis have significantly advanced the feasibility of Sobol' index computation for computationally expensive models and models with many inputs.
However, the limitations of existing methods still preclude their application to models with an extremely large number of inputs.
In this work, we present practical and theoretical extensions to the existing given-data Sobol' index method, which allow variance-based sensitivity analysis to be efficiently performed on large models such as neural networks, which have $>10^4$ inputs.
For models of this size, holding all input-output evaluations simultaneously in memory---as required by existing methods---can quickly become impractical.

Our extensions include a general definition of the given-data Sobol' index estimator with arbitrary partition, a streaming algorithm to process input-output samples in batches, and an asymptotic analysis of the new estimator that motivates a practical screening heuristic for small indices.
We show that the equiprobable partition employed in existing given-data methods can introduce significant bias into Sobol' index estimates even at large sample sizes and provide numerical analyses that demonstrate why this can occur.
We also show that the streaming algorithm can achieve comparable accuracy and runtime while substantially reducing memory requirements, enabling sensitivity analysis of models with much larger input dimension.
We demonstrate our novel developments on two application problems in neural network modeling.
\end{abstract}

\section{Introduction}
Variance-based sensitivity analysis (VBSA) is a powerful tool in uncertainty quantification, attributing fractions of model output variance to uncertainty in model inputs.
VBSA in the form of Sobol' indices has seen wide adoption across a range of problems~\cite{iooss_review_2015,razavi_future_2021}.
Sobol' first-order indices attribute the fraction of output variance to each input individually, while total-order indices attribute the fraction of output variance to each input as well as its interactions with other inputs.
Additionally, they are robust to nonlinear and nonmonotonic dependence of model outputs on inputs~\cite{iooss_review_2015}, making them well-suited for a range of scientific applications.

This paper develops methods for tractable computation and screening of Sobol' indices for application problems subject to the following challenges:
\begin{smitemize}
    \item Extremely high-cardinality input space (e.g., $10^4-10^5$).
    \item Model inputs cannot be prescribed; sensitivity analysis must be performed from unstructured input-output samples.
\end{smitemize}
These challenges, combined, lead to applicability gaps for existing Sobol' index methodologies, which we describe herein.

Traditional approaches to compute Sobol' indices are based on pick-freeze sampling designs, in which structured sample matrices are constructed to isolate the contribution of individual input variables to the output variance~\cite{saltelli_variance_2010,saltelli_making_2002,sobol_global_2001,sobol_sensitivity_1993,saltelli_global_2007}.
These pick-freeze methods require the ability to evaluate the model at prescribed input values associated with these structured sample matrices. 
Additionally, for a given sample size $N$, pick-freeze indices require $N(d+2)$ model evaluations, where $d$ is the number of inputs. 
For models with many inputs, this computational cost can quickly become prohibitive.
A variety of strategies have been proposed to mitigate this cost, including grouped sensitivity indices, which reduce the number of inputs down to the number of groups~\cite{lo_piano_uncertainty_2022,campolongo_effective_2007,b_ciuffo_sensitivity-analysis-based_2014,sheikholeslami_global_2019}, 
adaptive sampling strategies that iteratively allocate additional model evaluations to indices with high estimated uncertainty~\cite{lo_piano_variance-based_2021}, and improved Monte Carlo estimators that seek greater statistical efficiency for a fixed computational budget~\cite{azzini_sobol_2021}.
However, these strategies still rely on the ability to evaluate the model at prescribed inputs, which makes that unsuitable for the application problems of interest in this work.

Given-data approaches such as~\cite{borgonovo_common_2016,li_efficient_2016} seek to estimate Sobol' indices directly from unstructured input-output samples.
Compared to pick-freeze indices, these methods operate directly on $N$ samples and do not require additional structure sample evaluations.
Additionally, given-data approaches substantially broaden the applicability of VBSA by removing the requirement that model inputs are able to be prescribed. 
For example, such approaches enable VBSA for models with stochastic inputs, as well as for previously-performed sampling studies for which no additional model evaluations can be acquired.
However, existing methodologies are formulated as batch (or all-at-once) algorithms that assume it it possible to repeatedly access the full set of samples during estimation.
For the applications of interest here, with extremely large sample sizes and high-cardinality input spaces, the size of the complete dataset may exceed the memory available on conventional computing resources, motivating methods that avoid all-at-once computation.

A further consideration is the efficient screening of indices for these extremely high-cardinality input spaces, where a significant proportion of estimated indices may be negligible. 
A natural approach to this problem is to quantify estimator uncertainty and identify indices whose values are statistically indistinguishable from zero, e.g., through confidence intervals or hypothesis tests. 
Bootstrap procedures have been proposed to attain confidence intervals for both pick-freeze~\cite{archer_sensitivity_1997} and given-data estimators~\cite{borgonovo_common_2016}. 
However, bootstrap methods require repeated evaluations of the estimator over resampled datasets, making their application substantially more challenging for the large-scale problems considered here, in which the complete input-output sample set cannot be held in memory for multiple passes.
Another approach employs asymptotic confidence intervals and hypothesis tests based on the estimators' limiting distribution~\cite{gamboa_statistical_2016}; however, these developments were derived for classical pick-freeze estimators.

Collectively, the limitations of existing approaches described above motivate the methodological developments presented in this work. 
Building upon the given-data estimators of~\cite{borgonovo_common_2016,li_efficient_2016}, our key contributions are:
\begin{smitemize}
    \item A general given-data Sobol' index estimator applicable to any partitioning scheme, with accompanying empirical assessments of convergence properties for alternative partitions.
    \item Development of a novel algorithm to compute Sobol' indices in a streaming fashion as samples arrive.
    \item Asymptotic analysis of fixed-partition given-data Sobol' index estimators, extending existing consistency results and characterizing their convergence behavior.
    \item A practical screening heuristic for identifying indices that cannot be distinguished from zero-valued indices due to statistical noise, motivated by the asymptotic analysis.
\end{smitemize}
We demonstrate these novel capabilities on two application problems in analog neural networks, which were the motivation for this work.

The paper proceeds as follows: 
in~\cref{sec:background} we introduce Sobol' indices and given-data methods for their computation;
in~\cref{sec:methods} we define our generalized given-data Sobol' index estimator, our streaming algorithm for their computation, and our heuristic for filtering out small indices;
in~\cref{sec:results} we present a series of numerical studies investigating and demonstrating the efficacy of our novel methods;
in~\cref{sec:application} we present the application of our methods to two modeling problems in analog computing;
in~\cref{sec:conclusions} we discuss conclusions and future work.

\section{Background}\label{sec:background}
Sobol' indices are variance-based sensitivity measures that attribute the fraction of variance in a model output to input variables and their interactions. 
Their theoretical foundation in the Analysis of Variance (ANOVA) decomposition, as discussed in~\cite{sobol_sensitivity_1993,sobol_global_2001,prieur_variance-based_2017}.
Sobol' indices are an attractive option for sensitivity analysis because they measure sensitivity across the whole input space and because they are robust to nonlinear dependence as well as interaction effects between inputs. 

Here we focus on first-order (or main effect) Sobol' indices.
Let the model output of interest be denoted $f(\Xb)$ where $\Xb = [X_1, \ldots, X_d]$ are the uncertain inputs to the model.
Denoting all input parameters except $X_i$ as $\Xbi$, the first-order (or main effect) Sobol' index for input $X_i$ is defined as
\begin{align}
    S_i = \frac{
        \V_{X_i}\left( \E_{\Xbi}\left[f(\Xb) \given X_i \right] \right)
        }{\V(f(\Xb))}
        = 1 - \frac{\E_{X_i}\left[\V_{\Xbi}\left(f(\Xb)| X_i\right)\right]}{\V(f(\Xb))}. \label{eq:main_effect}
\end{align}
First-order indices measure the proportion of output variance attributed $X_i$'s variation alone, disregarding the effects of any interactions with other inputs on output variance.

The most common method for computing first- and total-order Sobol' indices, the pick-freeze method~\cite{saltelli_making_2002,prieur_variance-based_2017}, exhibits limitations that make it unsuitable for some practical applications. 
First, its cost can be prohibitive for computationally-expensive computational models with many inputs, since the cost to attain the full set of main and total effects indices for all inputs is $N(d+2)$, where $N$ is the number of independence samples used in the index estimators and $d$ is the number of inputs.
Second, pick-freeze methods require the ability to pass specific input values to the computational model, in order to evaluate the structured sample sets required for their computation.
This can be limiting in application problems where it is challenging or impossible to control model inputs, as is the case for our motivating application problem for analog neural network models.
This limitation in being able to control model inputs also precludes the use of grouped Sobol' indices, which have been shown to mitigate the computational costs associated with pick-freeze estimators by reducing the number of factors over which the indices are computed (i.e., reducing $d$)~\cite{lo_piano_uncertainty_2022,campolongo_effective_2007,b_ciuffo_sensitivity-analysis-based_2014,sheikholeslami_global_2019}.

Binning-based given-data methods~\cite{plischke_global_2013,borgonovo_common_2016,li_efficient_2016} address the computational challenges associated with pick-freeze methods.
As indicated by their name, they are constructed to operate on a given sample set without any specific structure.
Therefore, the number of model evaluations required for given-data methods is equal to the number of samples, $N$---crucially, the number of model evaluations required does not depend on the number of inputs.
While there has been some effort to develop given-data methods for total-order indices~\cite{zhai_space-partition_2014}, in practice they require far more samples to converge than first-order indices and thus to-date they have not been as practically feasible. 
For this reason, we focus on first-order indices in this work, and any future reference to Sobol' indices can be understood to refer to first-order indices.

Given-data methods use samples to estimate an inner statistic, then an outer statistic, by partitioning samples into $M$ bins over the input space.
The given-data procedure produces an estimate of the numerator of the first-order Sobol' index formulae in~\cref{eq:main_effect}, taking variance or expectation as the outer statistic, respectively.
Since given-data methods most often use the expectation as the outer statistic, we focus on approximations of the second equality of~\cref{eq:main_effect} in this work.
\section{Methods} \label{sec:methods}

We begin by defining the given-data first-order Sobol' index estimator.
For the rightmost definition of the first-order index in~\cref{eq:main_effect}, the given-data estimator approximates the numerator of the ratio:
$\E_{X_i}\left[\V_{\Xbi}(f(\Xb)\given X_i)\right]$. 
For simplicity of notation, subscripts on variance and expectation formulae will not be stated in further discussion unless needed for clarity.
The partition of the probability space for $X_i$ is defined as 
\begin{align}
\mathcal{A}_i = \bigcup_{k=1}^M A_k, \quad A_k \cap A_{j\neq k} = \varnothing, \quad A_k = [a_{k-1},a_k).
\end{align}
By the law of total expectation, 
\begin{align}
    \E\left[\V(f(\Xb)\given X_i)\right] = \sum_{k=1}^M \E\Big[\V(f(\Xb)\given X_i ) \;\Big|\; X_i \in A_k \Big] P(A_k),
\end{align}
where $P(A_k)$ is the probability of $X_i \in A_k$.
Given-data methods make the approximation
\begin{align}
    \E\Big[\V(f(\Xb)\given X_i ) \;\Big|\; X_i \in A_k \Big] \approx \V(f(\Xb)\given X_i \in A_k ). 
\end{align}
As discussed in~\cite{zhai_space-partition_2014}, as the partitioning of the space becomes infinitely fine (i.e.,~$M\rightarrow \infty,$ $P(A_k)\rightarrow 0$), this approximation becomes an equality. 
However, for finite $M$ this approximation may create numerical errors. 
We investigate the numerical errors arising from this approximation for finite partitions in~\cref{sec:partition_accuracy}.

For an equiprobable partitioning of the space, we arrive at the following expression for the numerator:
\begin{align*}
    \E\left[\V(f(\Xb)\given X_i)\right] 
    &\approx \sum_{k=1}^M \V(f(\Xb)\given X_i \in A_k ) P(A_k) \\
    &= \sum_{k=1}^M \V(f(\Xb)\given X_i \in A_k ) \left(\frac{1}{M}\right) 
    \approx \frac{1}{M} \sum_{k=1}^M s^2(X_i \in A_k).
\end{align*}
\purple{This equiprobable-partition form of the given-data Sobol' index estimator is the one that has historically been employed~\cite{plischke_global_2013,borgonovo_common_2016,li_efficient_2016}.}

For streaming algorithms it is challenging to enforce equiprobability; therefore, in this work we relax the assumption of an equiprobable partitioning. 
Our more general binned estimator for the first-order Sobol' index is computed as
\begin{align}
    \Shat = 1 - \frac{\EV}{\Vhat}, \quad\quad \EV = \sum_{k=1}^M s^2(X_i \in A_k) P_k, \quad\quad P_k = \frac{n_k}{N}, 
    \label{eq:general_estimator}
\end{align}
where $\Vhat=s^2$ is the sample variance, $P_k$ is the approximate probability of $A_k$, computed as the ratio of $n_k$, the number of samples in $A_k$, and $N$, the total number of samples.
Note that this estimator is very similar to one presented in~\cite{borgonovo_common_2016}.
However, that estimator was defined to approximate the left equality in the first-order index definition~\cref{eq:main_effect}, so that the outer statistic was a variance and the inner statistic was an expectation.

To address the challenges associated by our motivating application problems, which feature extremely high cardinality input spaces ($10^4-10^5$ inputs) so that all samples cannot be held in memory on conventional computing systems, we present the following extensions to the given-data method: in~\cref{sec:streaming_algorithm} we extend the algorithm to accommodate processing samples in a streaming or parallel fashion, in~\cref{sec:partitioning_schemes} we generalize the algorithm to accommodate partitioning schemes beyond equiprobable, and in~\cref{sec:heuristic_approach} we present theoretical results on the asymptotic behavior of the estimator for a fixed partition and a corresponding heuristic to filter out small indices that are statistically indistinguishable from a zero index.

\subsection{Given-data methods with streaming data}\label{sec:streaming_algorithm}
\begin{figure}[h!]
    \centering
    \includegraphics[width=0.9\textwidth]{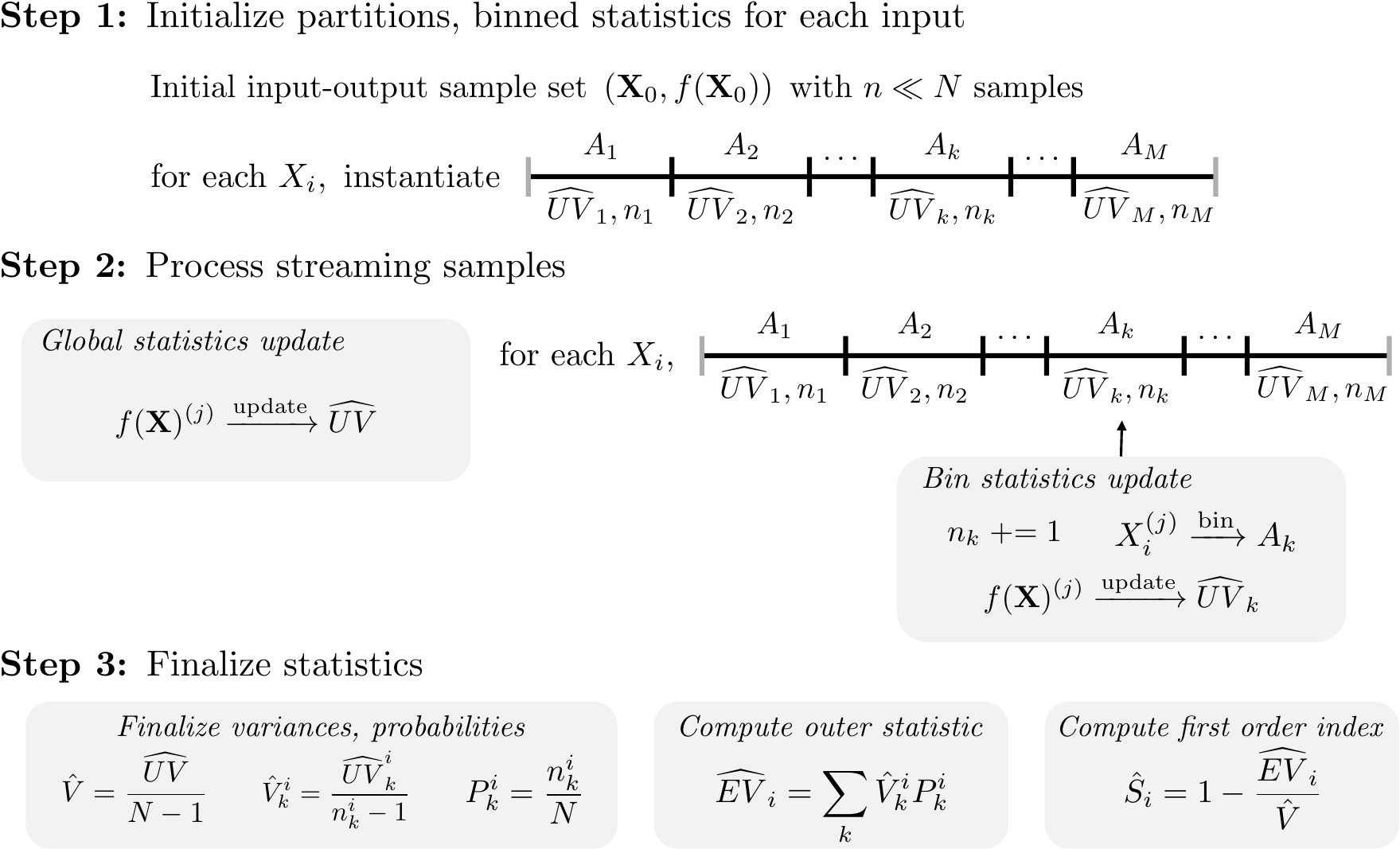}
    \caption{A depiction of the streaming given-data algorithm. $\UV$ denotes an unscaled sample variance, i.e., it has not been divided by the sample size. Steps 1 and 2 are applied for each input $X_i$; superscripts on statistics and bins are suppressed for notational simplicity.
    }
    \label{fig:streaming_algorithm}
\end{figure}

A streaming extension to the given-data method consists of three parts: (1) define a partition based on an initial sample set of size $n \ll N$, (2) update streaming statistics in each bin as samples arrive, and (3) finalize statistics once all samples have been processed.
The streaming given-data algorithm is detailed in~\cref{fig:streaming_algorithm}.

To update sample statistics in a streaming fashion we use the formulae presented in~\cite{chan_updating_1982,pebay_formulas_2008}.
Two sample sets $\Xb_1$, $\Xb_2$ of sample size $n_1$ and $n_2$, respectively, can be combined to compute the sample mean and variance for the combined sample set of size $n=n_1+n_2$. 
Denoting sample means for the sample sets as $\hat{\mu}_1$ and $\hat{\mu}_2$ and the unscaled variances (the sum of squares of differences from the current sample mean) as $\UV_1$ and $\UV_2$, the formulae to compute the combined sample mean and unscaled variance are:
\begin{equation}
    \begin{aligned}
        n &= n_1 + n_2, \qquad
        \delta &= \hat{\mu}_2 - \hat{\mu}_1, \qquad
        \hat{\mu} &= \hat{\mu}_1 + \frac{n_2}{n}\delta, \qquad
        \UV &= \UV_1 + \UV_2 +\frac{n_1n_2}{n} \delta^2.
    \end{aligned}
    \label{eq:streaming_update_formulae}
\end{equation}
These formulae are used to update the statistics in each bin as well as the total statistics. 

New samples are assigned to their appropriate bins for streaming updates using the edges between bins.
We use SciPy's binned statistic method to sort samples into bins, given the bin edges computed from the initial sample set.
Using this method we are able to compute binned statistics in batches and use the update formulas in~\cref{eq:streaming_update_formulae}, which allow for sample sets of any two sizes to be combined.

Once all samples have been processed, the statistics are finalized. 
For unscaled variances this means dividing by $n-1$ where $n$ is the number of samples used to compute $\UV$. 
Since a small initial sample is used to define the partition for each $X_i$, it is likely that the partition used in the streaming algorithm differs from the one that would be computed using all the samples at once. 
Because of this, even if an equiprobable partition was approximated with the initial sample set, it is likely that the final number of samples in each bin will not be exactly equal. 
This issue motivated our generalization of the given-data first-order Sobol' index estimator defined in~\cref{eq:general_estimator}, which relaxes the assumption of an equiprobable partition.

\subsection{Partitioning schemes}\label{sec:partitioning_schemes}

\begin{figure}[h]
    \centering
    \includegraphics[width=0.6\textwidth]{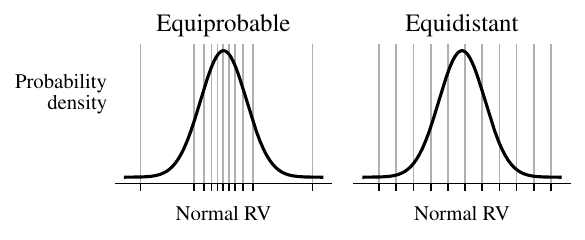}\vspace{-1em}
    \caption{Equiprobable and equidistant partitions defined for a normal random variable.}
    \label{fig:partition_illustration}
\end{figure}

The prevailing partitioning approach is an equiprobable partition: $M$ bins are identified, each with probability $1/M$. 
However, atypical input distributions such as continuous distributions with point masses (e.g., spike-and-slab densities) do not readily admit an equiprobable partition.
To address atypical distributions we introduce an ``equidistant'' partition, which instead uniformly discretizes over the range of $X_i$.
These two partitioning approaches are illustrated in~\cref{fig:partition_illustration}.
There are of course infinitely many ways an input space may be partitioned. 
More flexible partitioning methods are left as a subject of future research.

In both cases, computations will operate on an initial sample set for $X_i$ which we denote $\Xb_0^i$. 
Since $\Xb_0^i$ may not perfectly capture the range of $X_i$, interior bin edges are computed, and the outermost bin edges are assumed to be negative and positive infinity.
By setting the bounds to infinity, any new samples that fall outside the initial range of $\Xb_0^i$ will be assigned to the outermost bins. 
No matter the partitioning scheme, an important algorithmic choice that impacts accuracy is the number of bins, denoted $M$. 
The accuracy of the different partitioning schemes as a function of $M$ is investigated in~\cref{sec:partition_accuracy}.

\subsubsection{Equiprobable partition}\label{sec:equiprobable}
For a given $M$, an equiprobable partition equally divides the input space into bins with probability $1/M$. 
Denoting the cumulative distribution function (CDF) for $X_i$ as $F_{X_i}(x_i)$, the bin edges delineating the equiprobable partition are $B_k = F_{X_i}^{-1}(k/M),\;k=1,\ldots,M-1$.
We considered two approaches to estimate an equiprobable partition: a sample quantile approach and a kernel-density-estimate (KDE) based approach.
Both the KDE-based and sample quantile approaches employ an approximation of the CDF to estimate $B_k$.

The sample quantile approach computes $B_k$ using sample quantile formulae introduced in~\cite{hyndman_sample_1996} and implemented in NumPy. 
We employ the interpolated inverted CDF method in~\cite[Definition 4]{hyndman_sample_1996}, which estimates quantiles using a linear interpolation on the empirical CDF for $\Xb_0^i$. 
The KDE-based approach approximates the CDF using a Gaussian KDE approximation of the probability density for $X_i$, constructed using samples of $X_i$.
The KDE-based approach numerically integrates the KDE to approximate the CDF; the computational cost to evaluate the KDE on a fine grid to attain an accurate approximation thus makes the KDE approach significantly more costly than the sample quantile approach, which operates directly on $X_i$ samples.
However, since the KDE-based approach provides a smooth approximation of the CDF, it may provide a better approximation of $B_k$, especially for small sample sizes. 
To determine whether the additional cost of the KDE-based approach is warranted by yielding more accurate results, we compare the accuracy of both approaches in~\cref{sec:partition_initialization}.

\subsubsection{Equidistant partition}\label{sec:equidistant}
The equidistant partitioning approach is very simple.
Given initial samples $\Xb_0^i$, the bin edges delineating the equidistant partition are computed as $B_k = k \Delta x, \;i=1,\ldots,M-1$, where $\Delta x = (\max(X^i_0) - \min(X^i_0))/M$.

\subsection{Heuristic for filtering out small Sobol' indices} \label{sec:heuristic_approach}

\begin{figure}[h!]
    \centering
    \includegraphics{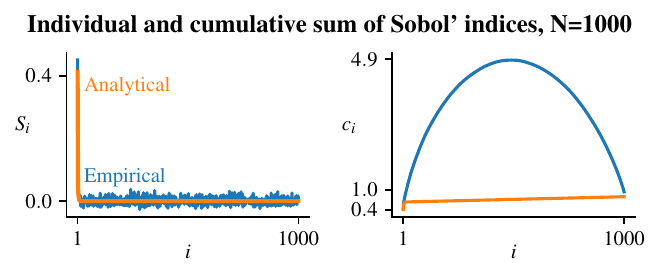}\vspace{-1em}
    \caption{Left: a comparison of empirical and analytical Sobol' indices for a model whose first 10 indices range from $\mathcal{O}(10^{-1})$--$\mathcal{O}(10^{-3})$, with the last 990 $\mathcal{O}(10^{-4})$.
    Right: A comparison of the cumulative sum of the indices, sorted maximum to minimum, denoted $c_i$.}
    \label{fig:why_filter}
\end{figure}

For applications with many inputs, the set of indices close to zero may have very high cardinality.
As shown in~\cref{fig:why_filter}, even if many inputs have indices near zero, statistical noise in their estimates can result in computed Sobol' indices that sum to far greater than the theoretical limit of $1$.
This noise can also make it challenging to distinguish a truly significant Sobol' index from a spurious one.
Since, in general, the aim is to identify a subset of parameters that are most important to the model output, it is attractive to filter out any computed indices that are indistinguishable from a zero Sobol' index at a given noise level (which decreases with sample size).

As discussed previously, existing methods to screen out near-zero indices based on confidence intervals or hypothesis tests are unavailable or challenging to implement for streaming, binned Sobol' index estimators. 
Therefore, we present a novel heuristic to filter out small indices in the case where many indices are near zero. 
The heuristic is underpinned by a new theoretical contribution of this work: an asymptotic characterization of the binned Sobol' index estimator that reveals qualitatively different limiting behavior for zero and nonzero indices. 
Exploiting this distinction leads naturally to the proposed screening criterion.

We first present the key theoretical results of this work. 
For readability, we defer their precise mathematical formulation and proofs to~\cref{app:theory} and summarize the principal conclusions here.
These results characterize the asymptotic behavior of the binned Sobol' index estimator and form the basis of the screening heuristic introduced in this section:

\textit{
\noindent$\mathbf{S_i>0.}$
Under the regularity conditions stated in~\cref{app:theory}, the binned estimator with fixed partition size $M$ for a nonzero Sobol' index is asymptotically normal with estimation error decaying at the rate $N^{-1/2}$.
}

\textit{
\noindent$\mathbf{S_i=0.}$
Under the same regularity conditions, the binned estimator for a zero-valued index is no longer asymptotically normal. 
Instead, its asymptotic distribution is a quadratic form in Gaussian random variables, with an additional positive shift arising from the use of unbiased sample variance estimators.
The estimator magnitude decays at the rate $N^{-1}$.
}

\noindent These results establish that the zero-valued index estimator has a distinct asymptotic sampling distribution, concentrating at rate $N^{-1}$ rather than $N^{-1/2}$ for nonzero indices. 
This separation motivates a screening criterion based on the variability of the zero-index distribution, which must be approximated in practice because the distribution is unknown.

We estimate the variability of the zero-index sampling distribution using the negative estimates in the observed ensemble, which are expected to arise predominantly from zero or near-zero indices in high-dimensional problems. 
These estimates are not direct draws from the zero-index sampling distribution, which is theoretically non-Gaussian and shifted to the right. 
However, the positive shift affects its location rather than its shape, and for practically relevant partition sizes, the distribution is expected to exhibit only modest positive skewness (see~\cref{rem:skewness}). 
We therefore use the symmetrized standard deviation of the negative Sobol' index estimates as an empirical measure of scale, denoted $\sigma$.

The validity of this approximation is investigated empirically in~\cref{sec:heuristic_assessment}, where we assess whether $\sigma$ provides a reliable estimate of the variability of the zero-index sampling distribution. 
We also examine the robustness of the proposed screening procedure to the choice of screening threshold.
We classify any estimated Sobol' index whose value is less than a prescribed multiple of $\sigma$ as statistically indistinguishable from zero. 
For a zero-mean Gaussian distribution, a $3\sigma$ threshold would already exclude approximately 99.7\% of the probability mass.
Since the asymptotic distribution of a zero-valued Sobol' index is theoretically positively skewed rather than Gaussian, we adopt the more conservative choice of $k=4$. 
This choice is heuristic rather than theoretically optimal, and the sensitivity of the screened results to this choice is examined in \cref{sec:heuristic_assessment}.
\section{Numerical investigations}\label{sec:results}
In this section we investigate the accuracy of the generalized Sobol' index estimator for all-at-once and streaming implementations, the accuracy of the considered partitioning approaches, and the efficacy and validity of our proposed heuristic for filtering out small indices.
Code to reproduce all numerical experiments in this section is publicly available at \url{https://github.com/sandialabs/MFUQ-Scalable-Given-Data-Sobol-Index-Estimators/}.

\subsection{Analytical test problems}\label{sec:test_problems}
For our numerical investigations we consider several test functions for which the Sobol' indices can be computed analytically.
Test problems are defined in full detail in~\cref{app:test_problems} and summarized here.
We consider the Sobol' G function~\cite{saltelli_variance_2010} and the Ishigami function~\cite{marrel_calculations_2009}.
Additionally, we consider the following polynomial function:
\begin{align}
    f_P(\Xb) = aX_1 + bX_2^2 + cX_1 X_2 + 0X_3, \label{eq:polynomial}
\end{align}
where $a,b,c$ are defined according to the random variable type to ensure nonnegligible first-order indices, and $X_3$ is a ``dummy variable'' included to assess zero index estimation.

\begin{figure}[h!]
    \centering
    \includegraphics{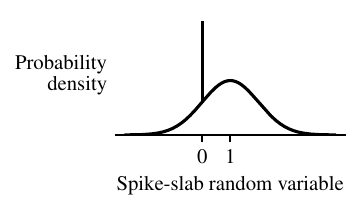}\vspace{-1em}
    \caption{PDF for the spike-slab random variable used in the polynomial test function.}
    \label{fig:spike_slab_pdf}
\end{figure}

To mimic the parameter distributions observed in our analog neural network application problems presented in~\cref{sec:application}, we also consider the polynomial function where inputs are ``spike-slab'' random variables, which have an almost-everywhere continuous probability distribution and a point-mass at one or more locations in the parameter space.
We define our spike-slab random variables here as the product of a Bernoulli and a normal random variable with probability $p$, yielding the probability density function (PDF)
\begin{align}
    f_X(x) = (1-p)\delta(x) + p f_Z(x).
\end{align}
The PDF of the spike-slab random variable is displayed in~\cref{fig:spike_slab_pdf}.

\subsection{Partition initialization}\label{sec:partition_initialization}
We investigate the effectiveness of the partitioning schemes for the streaming algorithm herein.
Of particular interest are (1) how the initial sample size and partitioning scheme impact the accuracy of the partition in the streaming algorithm and (2) the accuracy of the resulting Sobol' indices. 
For all investigations herein we generate 100 replicate samples to study the distribution of our results.
Our findings were consistent across bin counts of 50, 100, and 200, and for all test functions. 
We present only the 100-bin findings herein.
All results are provided in the code repository linked above.

\subsubsection{Bin count}
To investigate the accuracy of bin estimation as a function of the number of initial samples per bin, we perform the following study:
\begin{smitemize}
    \item For a given bin count $M$, we generate $ n_i M $ initial samples of our input variable $X$, where $n_i$ is the number of initial samples per bin.
    \item These initial samples are used to define bin edges with the chosen partitioning approach.
    \item We bin $1000M$ total samples according to the computed bin edges.
    \item We compare final bin counts to those achieved with an analytically defined partition.
\end{smitemize}
To understand the impact of the distribution of the random variable on accuracy, we investigate the effectiveness of the partitioning approaches for uniform, normal, exponential, and spike-slab random variables.

\begin{figure}[h!]
    \centering
    \includegraphics{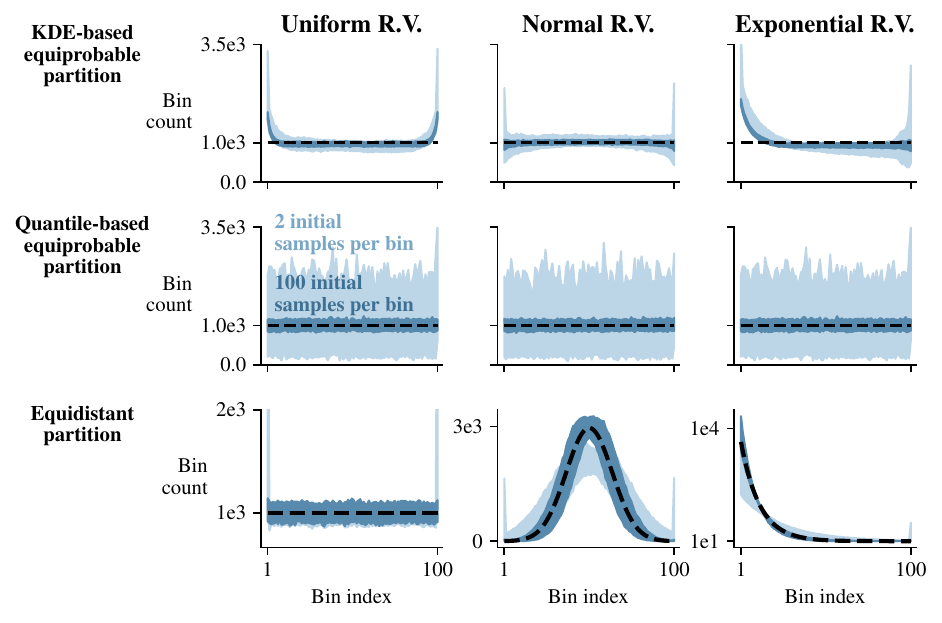}\vspace{-1em}
    \caption{Replicate bin counts for KDE and quantile equiprobable partitioning methods and the equidistant partitioning method. Shaded regions are the 5$^{th}$ and 95$^{th}$ percentiles over replicates, with lighter and darker shaded regions corresponding to 2 and 100 initial samples per bin, respectively. The black dashed line denotes the results of an analytical partition. For unbounded random variables the analytical equidistant partition was truncated at the points corresponding to a tail probability of $10^{-4}$.}
    \label{fig:bin_count_compare}
\end{figure}

We first compare the accuracy of the KDE and quantile approaches for equiprobable partitioning in the first two rows of~\cref{fig:bin_count_compare}. 
For all random variable types considered, the quantile approach exhibits significantly more variation in bin count for a small number of initial samples per bin relative to the KDE approach.
However, the KDE approach maintains significant error in bin count even for a large number of initial samples per bin for uniform and exponential random variables.
This behavior is likely due to the boundary bias of the Gaussian KDE for the bounded random variables.

Now we consider bin count accuracy using the equidistant approach in~\cref{fig:bin_count_compare}.
For all random variable types, the equidistant partitioning approach displays significant error in edge bin counts for low initial sample points per bin. 
This is likely due to the initial samples not adequately capturing the range of the random variable. 
With increased initial sample count, errors in edge bin counts diminishes.
The equidistant partitioning approach approximates the PDF of the random variable, with the quality of the approximation increasing as a function of the initial sample size per bin.
Further investigation would be needed to determine if this is an advantageous property.

\subsubsection{Dependence of Sobol' index accuracy on partition initialization}
To what extent does the accuracy of the first-order index computation depend on the accuracy of the partition estimation based on an initial sample set?
To investigate this question, we compare Sobol' indices computed with the streaming algorithm to Sobol' indices computed all-at-once (using all input samples to define the partition).
In these comparisons, we varied the number of initial samples and the random variable distributions.

\begin{figure}[h!]
    \centering
    \includegraphics{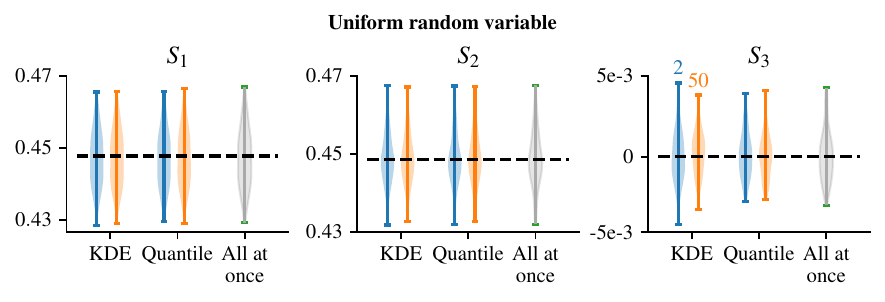}\vspace{-1em}
    \caption{Violin plots for KDE, quantile, and all-at-once equiprobable partitioning schemes. The bounds on the violin plots are the extrema over the replicate samples, while the shaded regions are the probability density over the replicates. The KDE and quantile violin plots show results with 2 and 50 initial samples per bin in blue and orange, respectively. The all-at-once approach partitions the entire sample set and therefore is not color coded. The analytical value of the Sobol' index for each parameter is shown as a black dashed line.} 
    \label{fig:equiprob_stream_vs_aao}
\end{figure}

The comparison between the KDE and quantile approaches for equiprobable partitioning and an all-at-once equiprobable partitioning is presented in~\cref{fig:equiprob_stream_vs_aao} for the polynomial test function with uniform random variables. 
The all-at-once partition uses all samples to define the partition. 
Results are shown for 100 bins with 100 samples per bin in total. 

For the equiprobable partition, the replicate index distributions are very similar across initialization choice (KDE and quantile) and initial sample size, indicating that these choices have minimal impact on index accuracy.
The insensitivity of the index estimators to initial sample size or KDE vs.~quantile equiprobable partitioning approach was consistent across random variable type, test function, and bin count.
This insensitivity also held for equidistant partitioning.
All other results are provided in the code repository linked above and are not presented here for brevity.

\subsubsection{Discussion}
While bin counts for different partition initialization schemes differed, it was found that these differences did not substantively impact Sobol' index accuracy for a fixed partition scheme.
There may be problems where these choices are influential, but we did not encounter any such cases in our investigations, across our suite of test functions, input distributions, and bin counts.
Given our current state of understanding, we provide the following guidance.

It was found that KDE and quantile approaches for equiprobable partitioning produced comparable accuracy in the index estimate.
This suggests that the quantile approach is advantageous for equiprobable partition initialization, since it achieves similar accuracy to the KDE approach for much lower computational cost.

Although increasing the number of initial samples improved bin edge estimates, no corresponding improvement in Sobol' index estimator accuracy was observed over all the considered test functions, input distributions, and initial sample sizes considered herein.
Even as few as two initial samples per bin produced results comparable to substantially larger initial sample sets, indicating that the index estimator is relatively insensitive to the accuracy of the initial partition.
This finding suggests the initial sample size may be chosen for convenience, e.g., using the batch sample size employed for streaming updates.

\subsection{Partition scheme accuracy comparisons}\label{sec:partition_accuracy}
Previous given-data estimators apply an equiprobable partition, perhaps due to its relationship to stratified sampling.
However, equipped with our generalized form of the estimator, we can ask: \textit{is an equiprobable partition the best choice?}
To answer this question we compare the accuracy of Sobol' indices computed with equiprobable and equidistant partitioning approaches for the polynomial test function.
For these analyses we processed all samples at once and used the KDE method to approximate the partition for the equiprobable case, since for spike-slab random variables the quantile method assigns two bin edges to the spike value, which causes numerical errors for the binned statistic computation.
The KDE method smooths the spike out, so the computation can proceed (acknowledging that the partition cannot be exactly equiprobable for this random variable type).

\begin{figure}[t!]
    \centering
    \includegraphics[width=\textwidth]{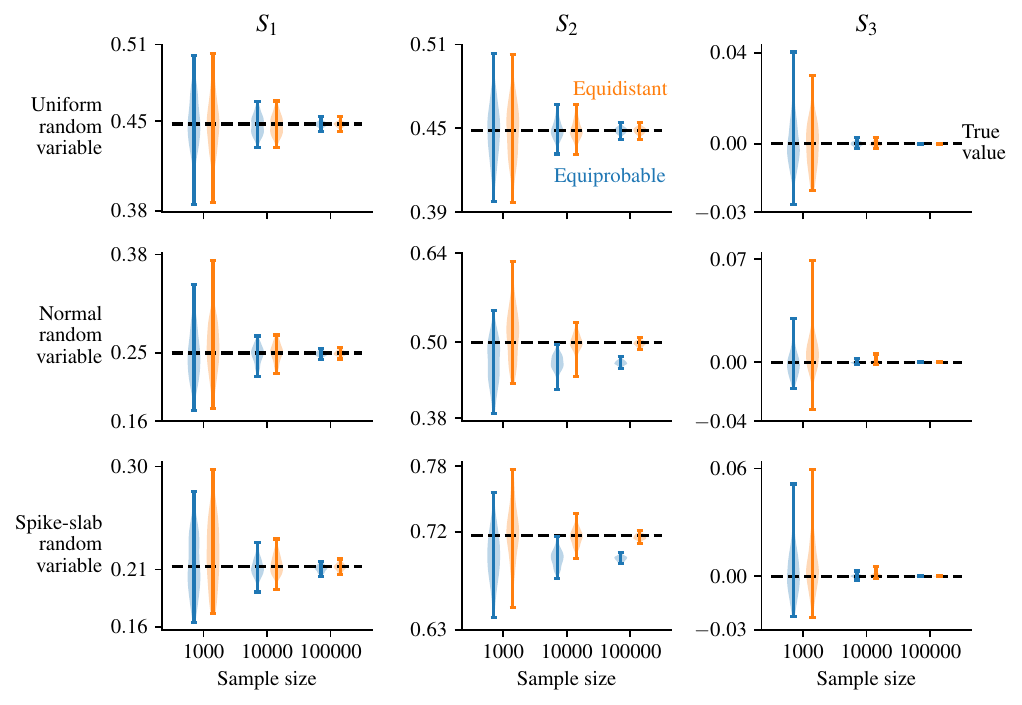}\vspace{-1em}
    \caption{Violin plots comparing replicate samples of Sobol' index estimates for equiprobable partitioning (blue) vs.~equidistant partitioning (orange), for 1000, 10000, and 100000 samples. True values are shown as a black horizontal line. All plots use 50 bins.}
    \label{fig:partition_compare}
\end{figure}

Replicate main effects indices computed using equiprobable and equidistant partitions are shown in~\cref{fig:partition_compare} for the polynomial test function, for uniform, normal, and spike-slab random variables.
For the uniform random variable case, the accuracy of the index estimator is comparable as a function of sample size for all inputs, which is expected since equiprobable and equidistant partitions are equivalent in this case.
However, for the normal and spike-slab random variables, accuracy of the approaches differs substantially for $S_2$. 
Even for 100000 samples, the equiprobable estimator for $S_2$ exhibits significant bias.

\begin{figure}[h!]
    \centering
    \includegraphics{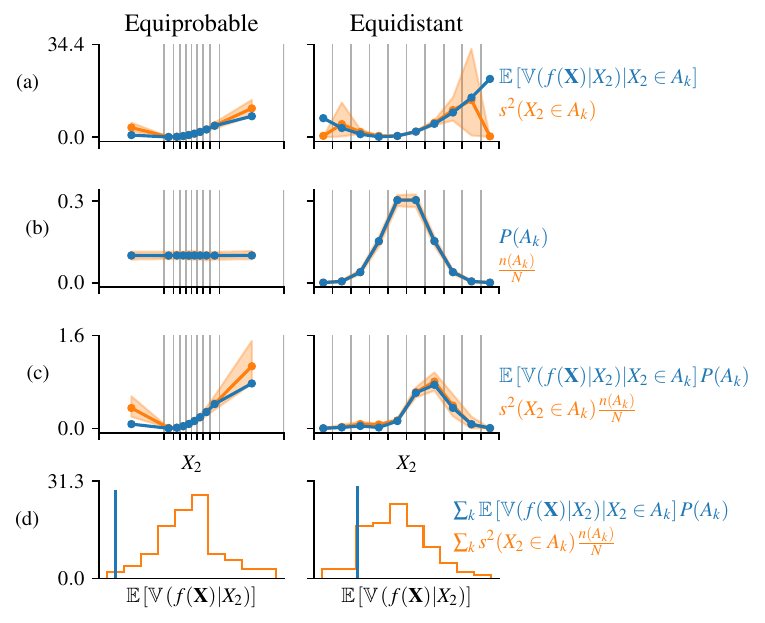}\vspace{-1em}
    \caption{A comparison of the estimators for equiprobable (left) vs.~equidistant (right) partitions of the normal-random-variable $S_2$ in~\cref{fig:partition_compare}. 
    \purple{Each row shares a $y$ axis. }
   High-accuracy estimates of exact statistics are shown in blue, while 100 replicates of the approximate statistics from 1000 samples are shown in orange. For (a)-(c) the shaded region indicates the 5$^{th}$ and 95$^{th}$ percentiles, and the solid line is the mean. In (d) the histogram of the numerator of~\cref{eq:main_effect} is shown for the 100 replicates vs.~the high-accuracy estimate.}
    \label{fig:partition_steps_compare}
\end{figure}
We illustrate the source of the bias in the equiprobable estimator of $S_2$ for the normal random variable in~\cref{fig:partition_compare} by comparing statistical estimates of the quantities in the equiprobable estimator vs.~highly-accurate estimates of the exact expressions in the expression of the numerator in~\cref{eq:main_effect}.
We compute $\E[\V(f(\Xb)\given X_i ) \;|\; X_i \in A_k]$ using SciPy's \texttt{expect} method, providing the analytical pointwise variance $\V(f(\Xb) \given X_i)$ and the bounds of $A_k$, $\left[a_{k-1},a_k\right)$.
Since we know the PDF and CDF of $X_i$, we know $P(A_k)$ exactly.

By investigating the steps in \cref{fig:partition_steps_compare} we see that for both partition types the approximate statistic in (a) is poor for some bins, but the equidistant estimator mitigates this error by assigning lower weight to those bins through its weighted average over bins.
In (a), both the equiprobable and equidistant partition poorly estimate the exact statistic at the edge of the domain (note that for the equidistant partition, there may be very few samples in the outermost bin, possibly zero or one sample, leading to zero variance). 
However, in (b), while the equiprobable partition assigns approximately equal probability to each bin, the equidistant partition approximates the PDF of the normal random variable, thus assigning low probability at the edges of the domain. 
Multiplying the quantities in (a) and (b) in (c), the discrepancy between the approximate and exact expressions persists for equiprobable, while they are mitigated for the equidistant case.
Taking the weighted average in (d), the equiprobable partition exhibits a more significant positive bias in the approximation of the numerator in~\cref{eq:main_effect} than the equidistant partition.
Since this estimator is divided by the global sample variance and then subtracted from 1 to compute the main effect index, this results in a negative bias in the main effect estimator, as observed in~\cref{fig:partition_compare}.

\begin{figure}[h!]
    \centering
    \includegraphics[scale=0.9]{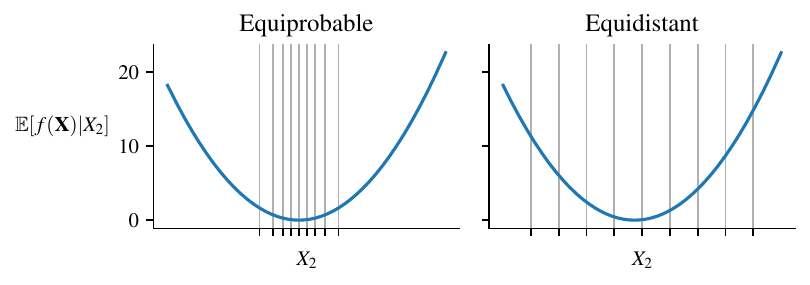}\vspace{-1em}
    \caption{Conditional variance for the polynomial test problem with respect to $X_2$ with equiprobable and equidistant partitions overlaid.}
    \label{fig:polynomial_E2_partition_compare}
\end{figure}

What causes the poor approximation in~\cref{fig:partition_steps_compare}(a)? 
By the law of total variance, 
\begin{align*}
\V(f(\Xb)\given X_i \in A_k) = \E[\V(f(\Xb) \given X_i) \given X_i \in A_k)] + \V\left( \E[f(\Xb)|X_i] \given X_i \in A_k \right). 
\end{align*}
Since the given-data estimators approximate $\E[\V(f(\Xb) \given X_i) \given X_i \in A_k)] \approx \V(f(\Xb)\given X_i \in A_k)$, the error will be large if $\V\left( \E[f(\Xb)|X_i] \given X_i \in A_k \right)$ is large.
Indeed, in~\cref{fig:polynomial_E2_partition_compare}, we see that $\E[f(\Xb)|X_i]$ varies most in the outer bins for both partitions, meaning this approximation is worst in these bins. 
Therefore, for a fixed number of bins $M$, the partition-induced bias is minimized by the partition that minimizes
\begin{align}
    \sum_{k=1}^M P_k \V\left( \E[f(\Xb)|X_i] \given X_i \in A_k \right).
    \label{eq:min_disc_err}
\end{align}
However, partition selection also affects the sampling variance of the estimator. 
Increasing the number of bins, or allocating disproportionately fine bins to regions of rapid variation in the conditional expectation, reduces the number of samples in those bins and increases statistical noise.
Therefore, partition selection exhibits a bias-variance trade-off.

This example illustrates that the equiprobable partition is not necessarily the optimal choice for all sensitivity analysis problems. 
\Cref{eq:min_disc_err} provides a general criterion for partition selection, but because the conditional expectation is problem-specific, no universally optimal partitioning scheme can be prescribed \textit{a priori}. 
When prior knowledge of the conditional expectation is available, it may be used to guide partition refinement while maintaining sufficient samples within each bin for reliable variance estimation.
In the absence of such knowledge, exploratory plots of $f(\Xb)$ against individual inputs may help identify regions where the conditional expectation changes rapidly.
If such regions coincide with relatively low-probability portions of the input space, equiprobable partitions may allocate excessively wide bins and drive up partition-induced bias, in which case an equidistant partition may be preferable. 
More generally, developing adaptive partitioning strategies that estimate the criterion in~\cref{eq:min_disc_err} from the available data and use it to guide partition refinement are an interesting direction for future work.

\subsection{Assessment of screening heuristic}\label{sec:heuristic_assessment}
\purple{
In this section, we assess the practical performance of the screening heuristic proposed in~\cref{sec:heuristic_approach}.
The heuristic is motivated by the theoretical results showing that the sampling distribution of a zero-valued Sobol' index has an $N^{-1}$ convergence rate and an asymptotic distribution given by a quadratic form in Gaussian random variables with an additional positive shift.
Although this limiting distribution may exhibit positive skewness, we expect that for practical problems with sufficiently large bin count, the skewness is sufficiently mild that a $k\sigma$ threshold provides a useful criterion for screening out statistically insignificant indices. 

To test the practical approximations associated with our proposed heuristic, we investigate the following questions:
\begin{smitemize}
    \item Is the sampling distribution of a zero-valued Sobol' index only mildly skewed in practice, even for highly skewed model outputs?
    \item Does the proposed standard deviation estimator based on the symmetrized negative index estimates accurately capture the variability of the zero-index sampling distribution, both in magnitude and in its predicted $N^{-1}$ convergence rate?
    \item Are the resulting screening decisions robust to the choice of threshold multiplier $k$?
\end{smitemize}
}

\subsubsection{Sampling distribution skewness}
We first investigate the shape of the sampling distribution of a zero-valued Sobol' index in practice to determine whether its non-Gaussianity is sufficiently mild that the proposed $\sigma$ approximation from negative index estimates provides a meaningful measure of sampling variability.
The theoretical analysis in~\cref{app:theory} shows that the zero-index distribution is asymptotically non-Gaussian with a positive shift.
We therefore examine empirically whether these features materially affect the use of a symmetric estimate of the sampling variability.

We expect the distribution to exhibit its greatest skewness for highly skewed output distributions and small sample sizes, where a few extreme output values may disproportionately influence the variance estimates appearing in the index estimator.
Intuitively, when relatively few samples are available within each bin, these extreme observations can have a substantial effect on individual bin variance estimates. 
As the sample size increases, however, the extreme observations are more evenly distributed across the bins, reducing their influence on any single estimate. 
We therefore anticipate that the sampling distribution becomes progressively less skewed with increasing sample size, approaching the asymptotic regime predicted by the theoretical analysis.

To investigate this behavior, we approximate the sampling distribution of a zero-valued Sobol' index by replicate sampling for three output distributions of increasing skewness over a range of sample sizes. 
The output is sampled from a Gamma distribution, $y\sim\Gamma(\alpha,\theta)$, where the variance is fixed at unity by setting $\theta=\alpha^{-1/2}$. 
We consider $\alpha=0.1, 0.01$, and $0.001$, with output skewnesses of $6.3$, $20$, and $63.2$, respectively. 
For each replicate the Sobol' index is computed for a dummy uniformly distributed input.

\begin{figure}[h!]
    \centering
    \includegraphics{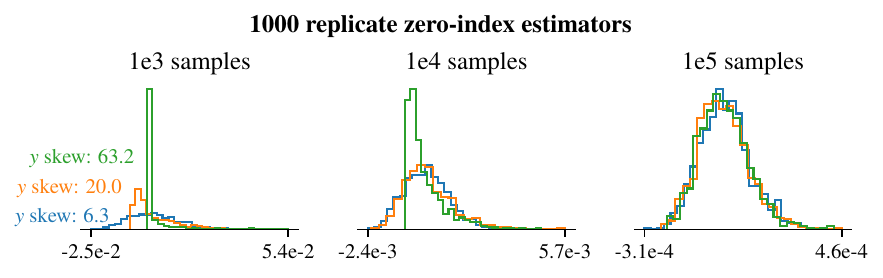}\vspace{-1em}
    \caption{Histograms of 1000 replicate zero-index Sobol' indices computed for outputs of increasing skewness while variance is held constant. The sample size used to compute the Sobol' indices increases from left to right from 1000 to 100000 samples.}
    \label{fig:gamma_noise_dists}
\end{figure}

As expected, the zero-index sampling distributions presented in~\cref{fig:gamma_noise_dists} show that positive skewness is greatest for the most highly skewed outputs and at the smallest sample sizes, and that skewness in the sampling distribution decreases with increasing sample size. 
It is worth noting that all three output distributions considered here are exceptionally skewed: a skewness exceeding $1$ is typically considered highly skewed, while the smallest considered here is $6.3$. 
Despite this, the sampling distribution for the 6.3-skewness case is already only mildly skewed for $N=10^3$, and even the 20-skewness case achieves similar mild skewness for $10^4$ samples.
Even the most extremely skewed case achieves mild skewness for $10^5$ samples.

These results suggest that, for all but the most extreme model outputs, a standard-deviation-based $k\sigma$ threshold provides a reasonable characterization of the sampling variability of zero-valued Sobol' indices, despite the non-Gaussianity and positive shift predicted by the asymptotic theory.
However, because highly skewed outputs can increase finite-sample asymmetry, we recommend computing output distribution skewness or examining output density estimates before applying the heuristic.

\subsubsection{Approximation of zero-index standard deviation}

\begin{figure}[h!]
    \centering
    \includegraphics{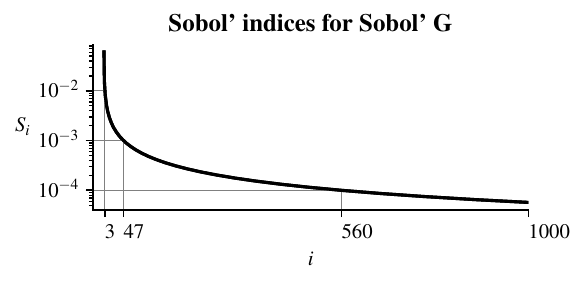}\vspace{-1em}
    \caption{Sobol' indices for the Sobol' G function defined in~\cref{eq:sobol_g} with 1000 input parameters. The number of indices falling above each relevant order or magnitude are shown in the $x$ axis.}
    \label{fig:sobol_g_indices}
\end{figure}

Having established that a standard-deviation-based threshold provides a reasonable characterization of the sampling variability of a zero-valued Sobol' index, we now investigate whether this standard deviation can be accurately estimated using a symmetrized standard deviation computed from negative indices.
We consider the 1000-dimensional Sobol' G function defined in~\cref{eq:sobol_g}. 
As shown in~\cref{fig:sobol_g_indices}, the function contains many near-zero first-order indices, making it well-suited for testing the proposed $\sigma$ estimator. 
The first-order indices explain only 35\% of the output variance, with the remainder arising from interactions.

The output distribution has estimated skewness of $13.4$, with a range of $11-16$ over ten independent estimates.
This falls well within the range examined in the previous subsection. 
We therefore expect the shape of the sampling distribution of zero-valued indices to remain only mildly skewed over the sampled sizes considered, ranging from $N=10^3$ to $N=10^5$. 

\begin{figure}[h!]
    \centering
    \includegraphics{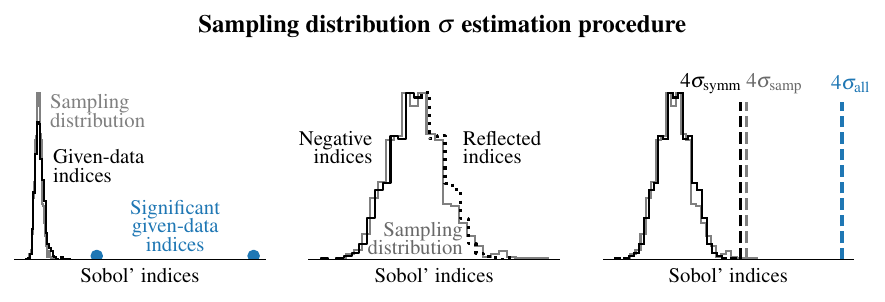}\vspace{-1em}
    \caption{
        Illustration of the proposed standard deviation estimator for the Sobol' G function using $N=10^4$ samples. 
        The sampling distribution of a zero-valued Sobol' index (obtained from 1000 dummy-variable replicates) is compared with the given-data indices, the symmetrized negative indices, and the resulting $4\sigma$ thresholds.
    }
    \label{fig:heuristic_procedure}
\end{figure}

Figure \cref{fig:heuristic_procedure} illustrates the proposed standard deviation estimator for a single realization of given-data indices with $N=10^4$. 
Two significant indices are highlighted in the left panel to distinguish them from the large collection of near-zero indices that overlap with the high-probability region of the sampling distribution.
The sampling distribution of a zero-valued Sobol' index is obtained using replicate samples for a dummy variable solely for validation purposes; in practice this distribution is unavailable. 
The proposed estimator reflects the negative indices about the origin and computes their standard deviation, $\sigma_{\rm symm}$. 
The resulting $4\sigma$ threshold closely matches that obtained from the replicate sampling distribution (denoted $4\sigma_{\rm samp}$) and falls on the upper bound of its high-probability region.
In contrast, computing the standard deviation over all indices ($\sigma_{\rm all}$) results in a substantially larger threshold because genuinely important indices inflate the estimate.

\begin{figure}[h!]
    \centering
    \includegraphics{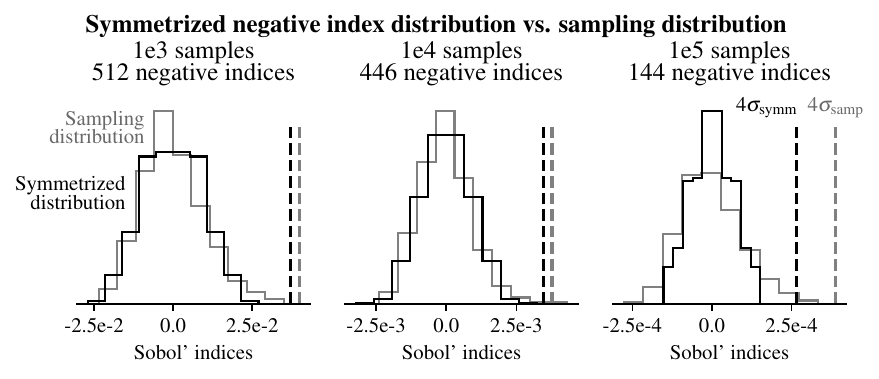}\vspace{-1em}
    \caption{
    Comparison of the sampling distribution of a zero-valued Sobol' index with the symmetrized negative given-data indices, together with the resulting $4\sigma$ thresholds. 
    Sample sizes increase from $10^3$ to $10^5$ left to right. 
    The number of negative indices used to estimate $\sigma_{\rm symm}$ is reported for each sample size.
    }
    \label{fig:heuristic_histogram_converge}
\end{figure}

Having illustrated the estimation procedure, we now examine how accurately $\sigma_{\rm symm}$ estimates the standard deviation of the zero-index sampling distribution as the sample size increases.
Figure~\cref{fig:heuristic_histogram_converge} compares the symmetrized negative indices with the true sampling distribution obtained from replicate calculations.
Across all sample sizes, the estimated threshold closely tracks the true threshold, consistently providing a slight underestimate.
As $N$ increases, fewer indices fluctuate below zero, reducing the number of samples available to estimate $\sigma_{\rm symm}$, and the approximation deteriorates somewhat.
However, since zero-valued indices converge at the faster rate $N^{-1}$ than nonzero indices at rate $N^{-1/2}$, the threshold becomes increasingly separated from genuine nonzero indices.
We therefore expect the practical impact of this modest underestimation to diminish as the sample size increases.

\begin{figure}[h!]
    \centering
    \includegraphics{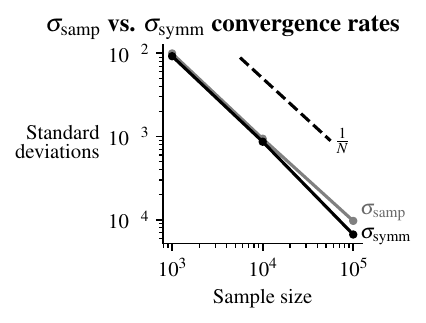}\vspace{-1em}
    \caption{Convergence rates for the sample standard deviations of the noise distribution of the Sobol' G function computed with 1000 replicates (grey) and of the symmetrized given-data index distribution (black).}
    \label{fig:heuristic_convergence_rate}
\end{figure}

As shown in~\cref{fig:heuristic_convergence_rate} $\sigma_{\rm samp}$ and $\sigma_{\rm symm}$ exhibit nearly identical convergence in this problem, confirming the theoretically predicted $N^{-1}$ scaling and providing further evidence that $\sigma_{\rm symm}$ accurately estimates the variability of the zero-index sampling distribution in practice, despite the positive shift predicted for the estimator.

\subsubsection{Robustness to threshold multiplier}

Finally, we investigate the sensitivity of the heuristic screening procedure to the choice of threshold multiplier $k$.
\Cref{tab:threshold_sensitivity} reports the sum of the retained index estimators for $k=0$, corresponding to the common practice of filtering out any negative estimates, and for one standard deviation on either side of our proposed $4 \sigma$ threshold. 
In line with intuition, increasing $k$ yields progressively more conservative screening by retaining fewer near-zero indices.
At the smallest sample size, $k=0$ the sum of retained indices sums to 4.37, far above a theoretical maximum of $1$ and despite the analytical sum of first-order indices being only $0.35$. 
This substantial overestimation arises from the accumulation of many small positive estimation errors across the 1000 inputs, illustrating the need for a positive screening threshold when many negligible indices are present.

As expected, increasing $k$ yields more conservative screening by removing a larger proportion of near-zero estimates. 
At smaller sample sizes the retained index sum depends significantly on $k$, reflecting the larger sampling variability in the estimated indices. 
As the sample size increases, however, the differences between the $k=3,4,$ and $5$ thresholds decreases substantially, with the retained sums approaching the analytical index sum of $0.35$.
These results suggest that the screening outcome becomes progressively less sensitive to moderate changes in the threshold multiplier as the sample size increases, with $k=4 $ providing a practical compromise between the more permissive $k=3$ and the more conservative $k=5$.
\begin{table}[h!]
    \centering
    \begin{tabular}{l|rrrr}

$N$ & $k=0$ & $k=3$ & $k=4$ & $k=5$ \\
\hline
1000 & 4.37 & 0.52 & 0.18 & 0.10 \\
10000 & 0.68 & 0.23 & 0.15 & 0.12 \\
100000 & 0.36 & 0.31 & 0.29 & 0.27 \\

\end{tabular}

    \caption{Sum of retained indices for different threshold multipliers and sample sizes.}
    \label{tab:threshold_sensitivity}
\end{table}
\section{Application to analog neural networks}\label{sec:application}

\begin{figure}[h!]
    \vspace{-1em }
    \centering
    \includegraphics[scale=1.0]{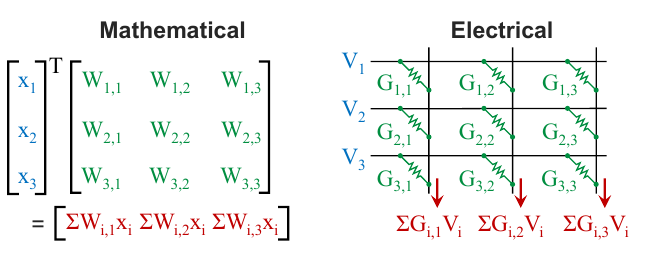}\vspace{-1em}
    \caption{Left: Matrix-vector multiplication $Wx=y$ (input vector moved to the left and transposed for clarity). Right: Implementation of the matrix vector multiplication in the analog electrical domain using circuit laws within a resistive memory array. Image from \cite{CrossSim}.}
    \label{fig:mat_vec}
\end{figure}

Analog in-memory computing hardware has the potential to perform matrix-vector multiplications at a much lower energy cost than conventional digital architectures, making it an attractive option for edge-computing applications where power constraints require high energy efficiency, such as mobile devices and satellites.
Analog computing uses Ohm's and Kirchoff's laws to implement matrix-vector multiplications by applying voltages to the rows of a cross-bar array of resistive memory devices (each with a programmable conductance) and measuring the resulting current that flows along the columns, as illustrated in \cref{fig:mat_vec} \cite{CrossSim}. 
A collection of such cross-bar arrays (with supporting digital components) can efficiently process neural network inference, where each array stores a matrix of trained weights \cite{Xiao:2020}.

Although efficient, this analog hardware implementation comes with multiple sources of uncertainty and approximation error.
One source of approximation error is the device programming error, which refers to the precision with which the conductances in the network are programmed to match the digital weights.
Higher conductance precision requires more steps in the iterative fine-tuning of device conductances, and therefore requires more energy for hardware configuration.
Other sources of uncertainty include read noise (in which thermal noise and flicker noise cause a conductance to have a different value each time the analog network is evaluated), and drift in the conductances over time.

The goal of this work is to identify which conductances should be implemented with higher precision to improve overall network predictive performance as energy efficiently as possible.
As such, we focus on the programming error for the analog network conductances, and use Sobol' indices to identify the most important conductances to network output accuracy.
While read noise (and other sources of uncertainty and error) are neglected in this analysis, we postulate that the conductances that have the most impact for their programming errors will also have a similar relevance for the other errors.
We consider two examples: a satellite detection network with $\geq 10^4$ parameters and a classification network with $\geq 10^5$ parameters.

While total-order indices would, in principle, provide a more complete measure of conductance importance by accounting for interaction effects, their estimation requires pick-freeze sampling. 
This is not possible in our setting because the conductances are not directly controllable in either the physical hardware or the hardware simulator; only stochastic realizations of the programmed network can be evaluated. 
Therefore, the resulting ranking is not guaranteed to identify the optimal set of conductances to target for maximal programming precision. 
Nevertheless, since first-order indices are lower bounds on the corresponding total-order indices, prioritizing conductances with the largest main effects is expected to reduce output variability, even when interaction effects are significant.

\subsection{Satellite Detection Network}\label{sec:sat}
The first application is a neural network to detect the occurrence of point-like events among clutter in synthetic satellite imagery~\cite{Xiao:2023}.
This is representative of a typical edge-computing application where it is desirable to first analyze images onboard a satellite before deciding to send only the most salient data to Earth over a slow datalink for further processing.
\begin{figure}[h!]
    \centering
    \includegraphics[scale=0.4]{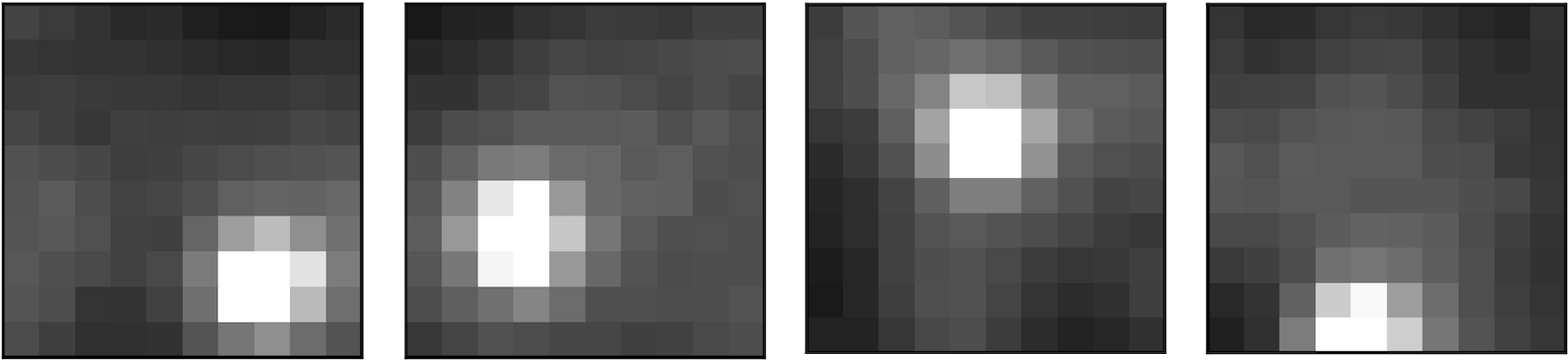}
    \caption{Example input images for the satellite detection network.}
    \label{fig:sat_input_imgs}
\end{figure}
The inputs for this network consist of grayscale images with 10 by 10 pixels, as illustrated in~\cref{fig:sat_input_imgs}. 
In~\cite{Xiao:2023} the network was formulated and trained to determine whether an input image contains an event (a white spot), the peak amplitude of the event (the brightness of the spot), and the x-y location of the peak amplitude.

An analog hardware implementation of this neural network is simulated using the CrossSim Python-based crossbar simulator, which can be used to simulate the behavior of a fabricated analog computational arrays~\cite{CrossSim}. 
The network is composed of two convolutional and two densely connected layers, resulting in 10696 weights for which we will compute Sobol' indices.
Further architectural details are provided in~\cref{app:satnet_architecture}.
Following~\cite{Xiao:2023}, this work uses the programming error model of a SONOS flash memory device, where the error follows a Gaussian distribution but with a state-dependent standard deviation~\cite{Agrawal:2022}.

\begin{figure}[h!]
    \centering
    \includegraphics[scale=0.4]{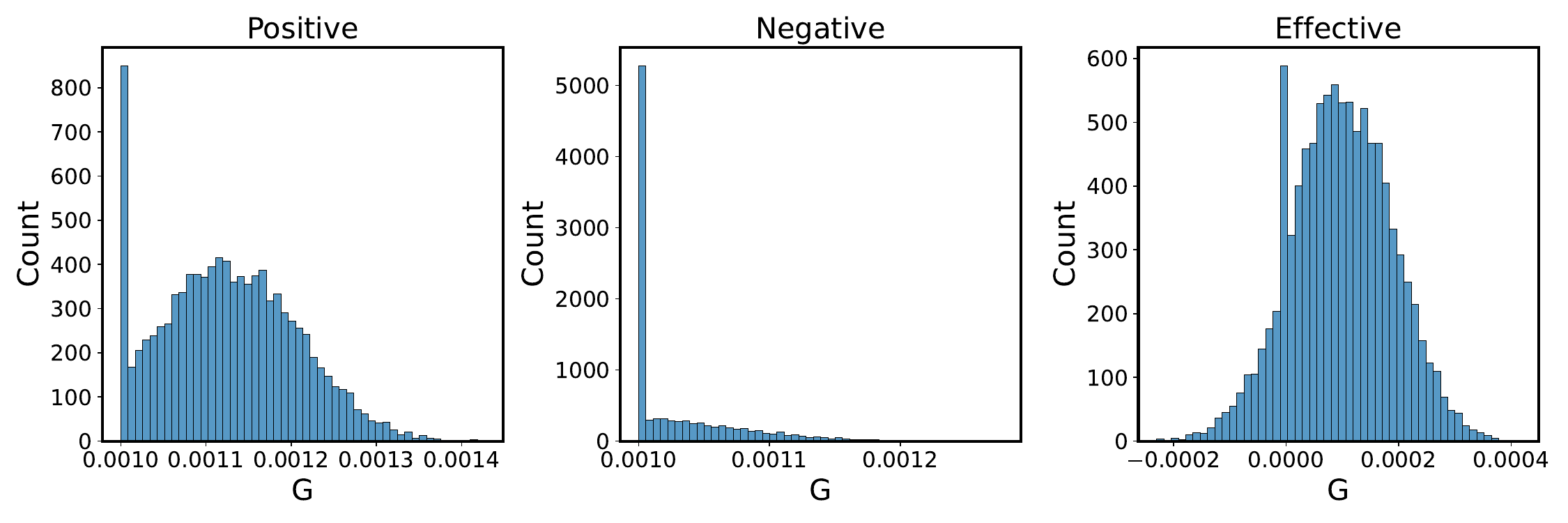} \vspace{-1em}
    \caption{
        Distribution of the effective conductance representing weight (14,72) in layer 2 of the satellite detection network. The threshold positive (left) and negative (middle) conductances produce a spike-slab distribution for the effective conductance (right). Conductance values are normalized by the maximum conductance in the network.
    }
    \label{fig:conductance_spikes} 
\end{figure}

An inherent property of analog in-memory computing hardware is that physical conductances can only be positive. 
Therefore, weights with negative values are represented using an \emph{effective conductance} $G_{\textrm{eff}} = G_+ - G_-$, obtained by subtracting the currents through two positive conductances.
Since both $G_+$ and $G_-$ are truncated below a threshold, with all smaller values mapped to the threshold conductance, many samples of $G_{\textrm{eff}}$ are exactly zero.
As a result, the effective conductances for many network weights exhibit a spike-slab distribution, as illustrated in~\cref{fig:conductance_spikes}.
We employ the equidistant partition scheme herein since it appeared to be more robust to such situations in~\cref{sec:partition_accuracy}

We are interested in how sensitive the network outputs are to the inaccuracies resulting from an analog resistive memory implementation of the weights in the neural network. 
The output for which we compute Sobol' indices is the average predicted peak amplitude over a test set of 5000 images. 
We compute a Sobol' index for each of the 10696 weights in the analog network.
A total of 50000 randomly sampled conductance sets and their corresponding average predicted peak amplitudes were generated.
The distribution has skewness $-0.34$, indicating it falls within the regime for which the screening heuristic developed in~\cref{sec:heuristic_approach} and evaluated in~\cref{sec:heuristic_assessment} is expected to perform well.

\begin{figure}[h!]
    \centering
    \includegraphics{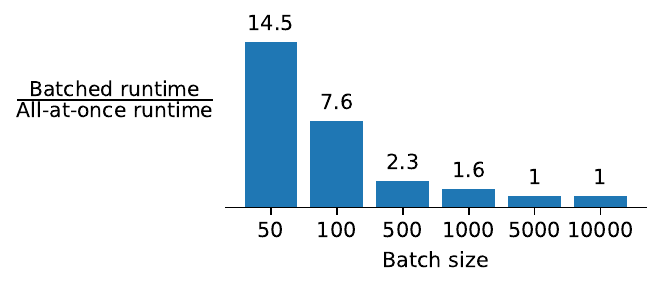} \vspace{-1em}
    \caption{Run time ratio of the streaming algorithms using different sized batches of samples compared to processing all samples in one batch.}
    \label{fig:sat_batch_performance} 
\end{figure}

Given the moderate size of this network, it is still possible to store and process all 50000 samples at once on a laptop with a moderate amount of memory. 
This allows us to study the computational performance of the streaming algorithm as a function of batch size.
In~\cref{fig:sat_batch_performance} we show the relative computational cost of computing the Sobol' indices with respect to each of the 10696 inputs using batches ranging from 50 to 10000 samples, compared to processing all 50000 samples at once.
Processing all 50000 samples at once took about 66 seconds on an Apple Macbook Pro with the M3 Max chip and 36GB of RAM using Python 3.12.4.

Working with small batches clearly slows down the algorithm, likely due to lack of vectorization in the streaming statistics update---we loop over batches of samples to update the binned and total statistics using the formulas defined in~\cref{eq:streaming_update_formulae}. 
We hypothesize that larger batches benefit from vectorization in SciPy's binned statistic method, which is used to compute the binned variance for the incoming batch of samples.
For batch sizes starting at 1000, the relative runtime is comparable to all-at-once processing of the samples, being within a factor of 2. 
For batch sizes 5000 and 10000, parity in runtime is achieved.

\begin{figure}[h]
    \centering
    \includegraphics[width=0.85\textwidth]{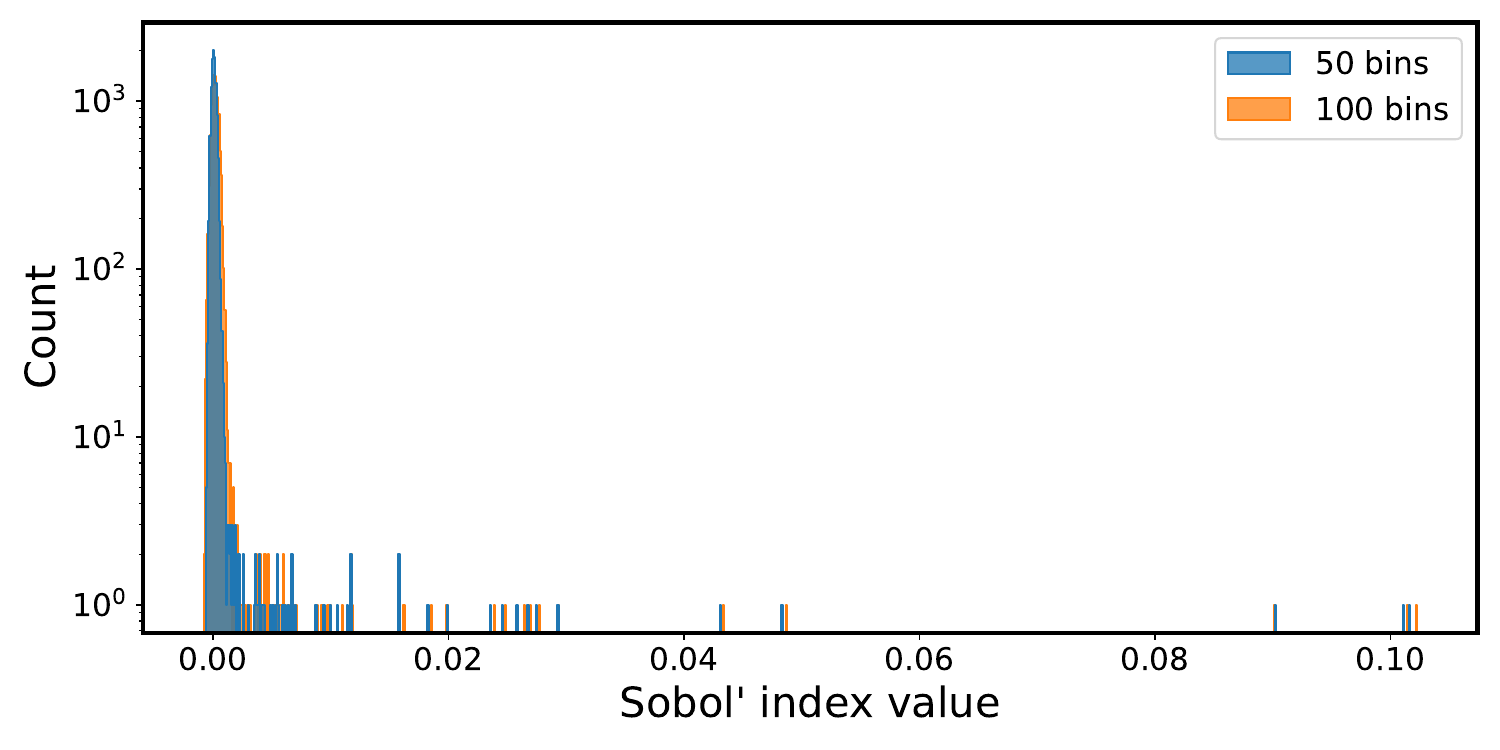} \vspace{-1em}
    \caption{Histogram of the computed main Sobol' indices for the satellite network w.r.t. each of its 10696 inputs computed with both 50 and 100 bins. While some of the Sobol' indices are as large as 0.1, most Sobol' indices are very small. Computing the Sobol' indices with 100 bins results in more noise than when using 50 bins.}
    \label{fig:sat_sobol_hist} 
\end{figure}

Given the large number of inputs, \cref{fig:sat_sobol_hist} summarizes the computed Sobol' indices using a histogram.
Many of the Sobol' indices are very small, which is expected given the large number of inputs.
Moreover, the nonlinear network may exhibit substantial interaction effects, further reducing the magnitude of the first-order indices computed here.
While a few indices are as large as 0.1, many are clustered around zero, with both positive and negative estimates arising from statistical noise, as observed for the Sobol' G function in~\cref{sec:heuristic_assessment}.
The indices computed with 100 bins exhibit greater noise for
The estimates computed with 100 bins exhibit greater variability among the near-zero indices, consistent with increased statistical noise; therefore, we use the 50-bin results in the remainder of the analysis.
However, we note that the largest indices were consistent between the 50- and 100-bin results.

We employ the heuristic discussed in~\cref{sec:heuristic_approach,sec:heuristic_assessment} to screen out indices that are statistically indistinguishable from a zero index. 
For the current analysis with 50 bins, the noise threshold evaluates to $6.5 \times 10^{-4}$.
Using the default threshold multiplier $k=4$, 209 of the 10696 weights are classified as relevant, with a retained index sum of $0.96$. 
Varying the threshold multiplier from k=3 to k=5 changes the retained sum only moderately, from $1.14$ to $0.90$, indicating that the screening result is reasonably insensitive to the precise threshold choice.
Since the retained indices remain affected by statistical noise, however, these sums should not be interpreted as accurate estimates of the total first-order variance explained.

Regardless of the remaining noise in the set of relevant Sobol' indices, a set of 209 likely relevant neural network weights is much more manageable than a set of 10696. 
Encoding these 209 weights with less variability should reduce the variability in the average predicted peak amplitude over the test set considerably. 
Verification of the actual percent variance attributed to these weights is the subject of future work.
\begin{figure}[h!]
    \centering
    \includegraphics[width=0.9\textwidth]{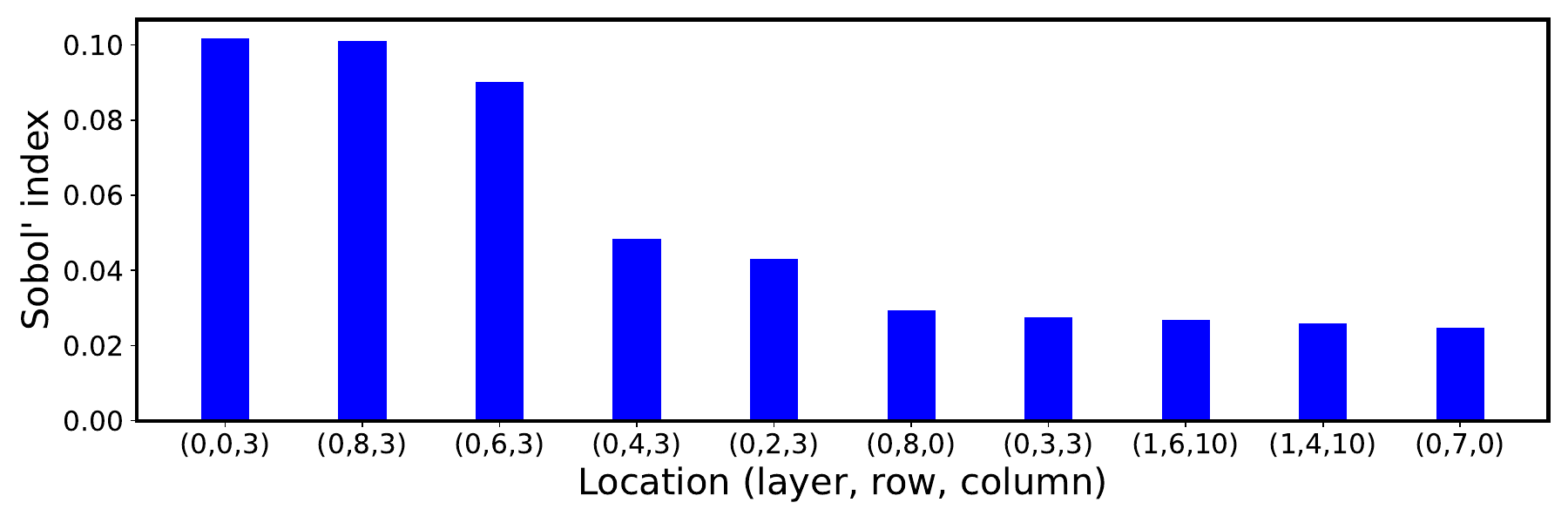} \vspace{-1em}
    \caption{Bar graph of the 10 largest Sobol' indices for the satellite network.}
    \label{fig:sat_S_largest} 
\end{figure}
To further illustrate this point, \cref{fig:sat_S_largest} shows the 10 largest Sobol' indices along with their location in the network. 
These 10 Sobol' indices sum to 0.52, suggesting that a relatively small subset of weights dominates the first-order sensitivity.
Of these values, 8 correspond to weights that are present in layer 0, indicating that this layer has a particularly strong influence on the network predictions.

\subsection{CIFAR-10 Image Classification Network}\label{sec:cifar10}

The second application we consider here is a neural network to perform classification of the CIFAR-10 dataset~\cite{Krizhevsky2009}. 
This dataset consists of $32 \times 32$ pixel color images that belong to one of 10 classes: airplane, automobile, bird, cat, deer, dog, frog, horse, ship, and truck. 
To classify these images, we use a pre-trained 14-layer Residual Network based on~\cite{He:2016}, available on Github at~\cite{CrossSimModels}.
This ResNet-14 network features 174128 weights for which we will compute Sobol' indices.
Further details about its architecture are presented in~\cref{sec:cifar_architecture}.

Our quantity of interest is the fraction of 1000 test images that were correctly classified for a given sample of the conductances. 
In the absence of programming errors (i.e., using the nominal values of the conductances), the ResNet-14 model achieves a classification accuracy of 88.9\%.
As in~\cref{sec:sat}, we use CrossSim to simulate the behavior of an analog hardware implementation of this network, employing the same effective conductance approach as in~\cref{sec:sat}. 
For this network, we used a state-independent programming error model~\cite{Xiao:2022}, which applies a normally distributed error to each conductance, with a standard deviation $\Delta G = \alpha G_{\text{Max}}$, with $\alpha = 0.025$, 
which reduces the average classification accuracy to approximately 68\%.
We therefore seek to identify the effective conductances whose programming errors have the largest first-order influence on classification accuracy.
The distribution of the classification accuracy (not shown) is smooth with skewness $-1.2$, suggesting that the screening heuristic is applicable to this example.

For this network, with about 175K inputs, the amount of RAM required to hold all samples in memory grows very quickly with the number of samples. 
For example, a batch of $10^4$ samples stored at 64 bits per conductance requires $\sim$13GB of memory.
As such, the only way to make the GSA computationally feasible is to employ the streaming algorithm.
Based on past numerical experiments, we use 50 bins for the given-data estimator, and we process the samples in batches of $10^4$ for runtime efficiency.

\begin{figure}[h!]
    \centering
    \includegraphics[width=0.85\textwidth]{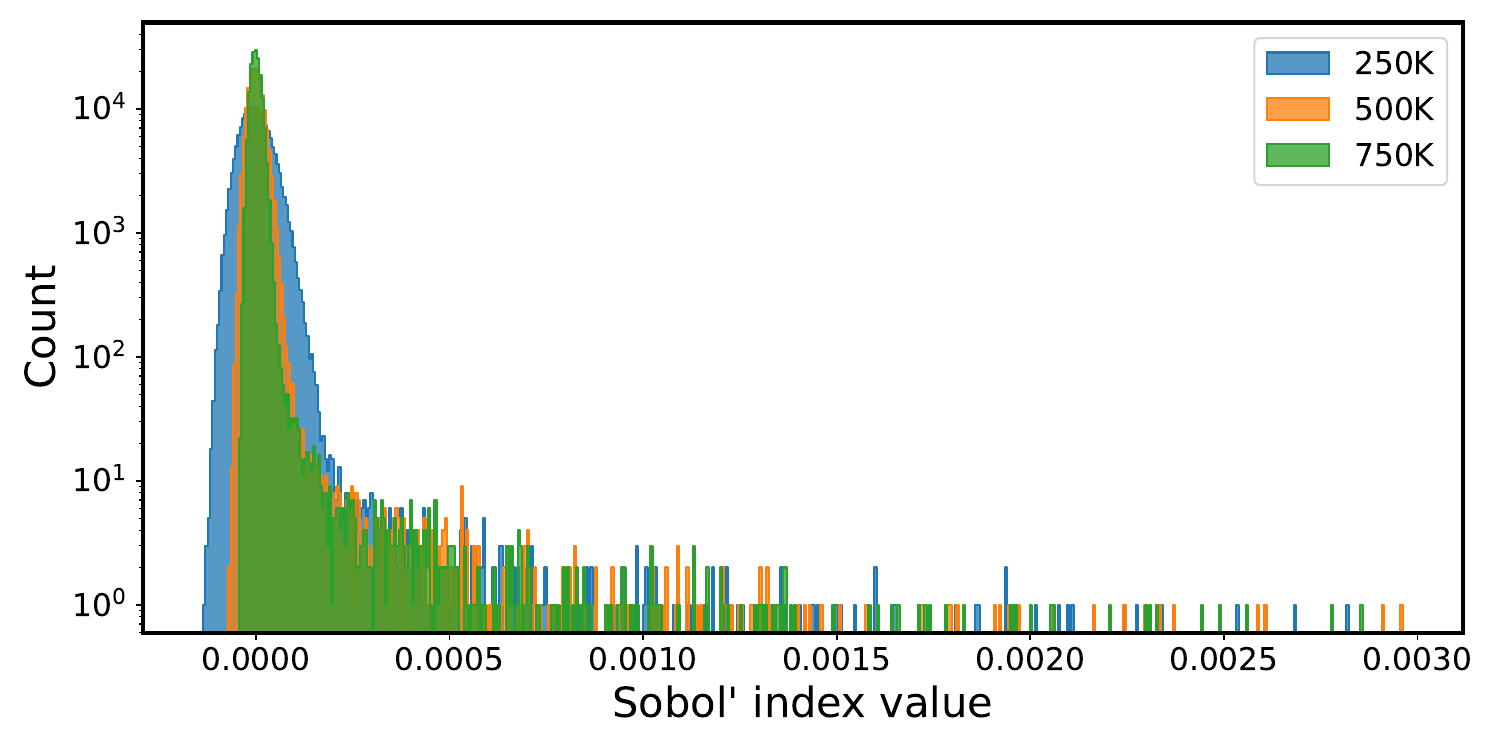} \vspace{-1em}
    \caption{CIFAR-10 Sobol' index histograms for increasing sample size.}
    \label{fig:cifar10_hists} 
\end{figure}
\begin{figure}[h!]
    \centering
    \includegraphics[scale=0.9]{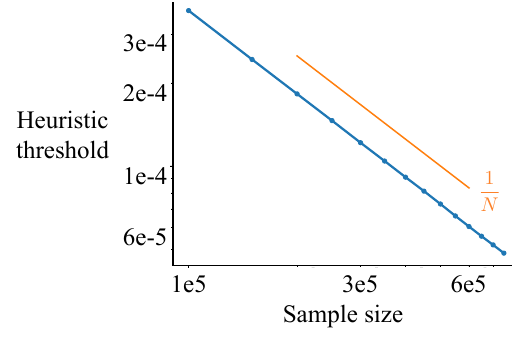} \vspace{-1em}
    \caption{Screening threshold versus sample size for the CIFAR-10 analysis.}
    \label{fig:cifar10_noiseconv} 
\end{figure}

To assess the convergence of the resulting Sobol' indices with the number of samples, we save snapshots of the running statistics every $5\times10^4$ samples, up to $7.5\times10^5$ total samples.
The Sobol' indices for increasing sample sizes are shown in~\cref{fig:cifar10_hists}.
As in the satellite case, there is a large cluster of Sobol' indices near 0, with a long tail towards larger values. 
Similar to~\cref{sec:heuristic_assessment}, the distribution of indices about zero contracts with increasing sample size.
Furthermore, the screening threshold shown in~\cref{fig:cifar10_noiseconv} decreases at the expected $N^{-1}$ rate, consistent with the asymptotic analysis in~\cref{app:theory}.

In the results with 750K samples, this noise cut-off is $4.84 \times 10^{-5}$, and 1205 Sobol' indices exceed the threshold.
Using the default threshold multiplier $k=4$, the sum of retained indices is 0.28.
Threshold multipliers of $k=3$ and $k=5$ yield sums of 0.34 and 0.26, respectively, indicating that the screening result is relatively insensitive to the precise threshold choice.
As is expected for this highly nonlinear network, substantial interaction effects are likely present, so the first-order indices are not expected to explain the majority of the output variance.
Nevertheless, the analysis identifies fewer than 1\% of the more than 175K conductances as having significant first-order influence, providing a substantial reduction in the set of weights to prioritize for higher-precision programming.

\begin{figure}[h!]
    \centering
    \includegraphics[width=0.9\textwidth]{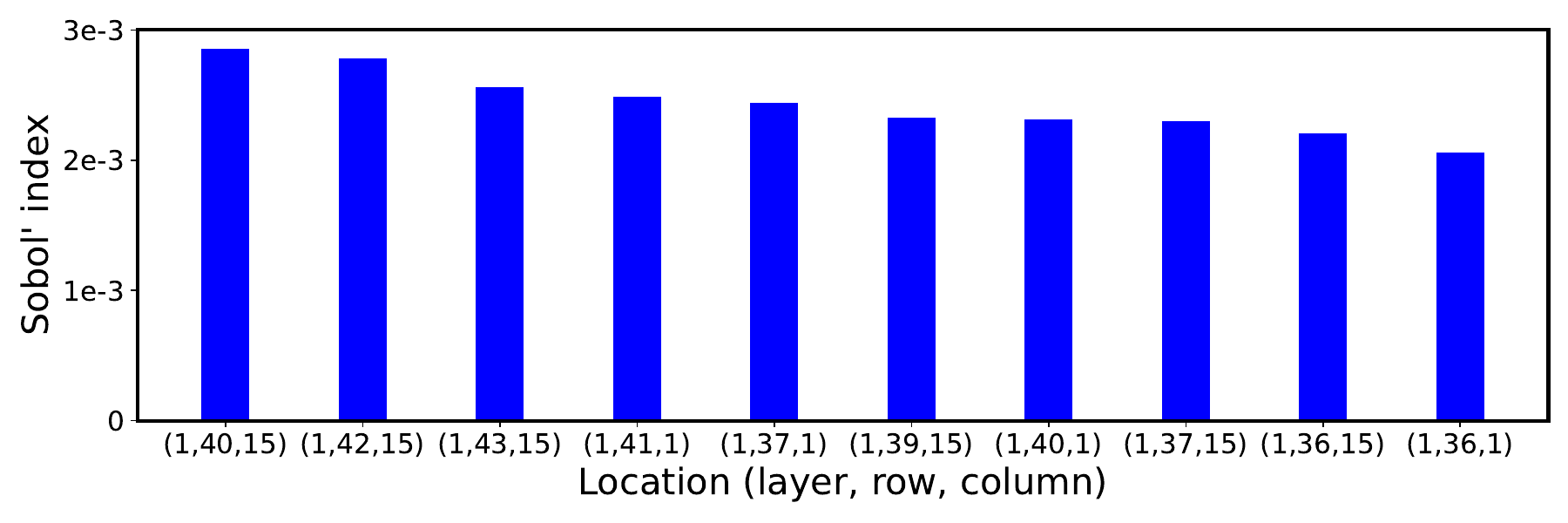} \vspace{-1em}
    \caption{Bar graph of the 10 largest Sobol' indices for the CIFAR-10 network.}
    \label{fig:cifar10_s_largest}
\end{figure}
As shown in~\cref{fig:cifar10_s_largest}, the 10 largest Sobol' indices correspond to conductances in layer 1. 
Furthermore, 768 of the 1205 retained indices are located in this layer, and their sum is 0.24, compared to the 0.28 across all retained indices.
Thus, the vast majority of the retained first-order sensitivity is concentrated in layer 1.
Note that in the satellite example, the most relevant Sobol' indices corresponded to conductances in layer 0, the first convolutional layer to process the input images. 
In the current ResNet architecture, the most influential layer (1) is the second layer to process the images, but the first layer in a block with a ResNet shortcut (see \cref{fig:resnetarch}).
This non-intuitive result illustrates the value of sensitivity analysis for identifying which weights to prioritize for higher-precision programming. 
\section{Conclusions}\label{sec:conclusions}

In this work we developed several practical and theoretical extensions to given-data methods for computing first-order  Sobol' indices, motivated by sensitivity analysis of analog neural networks, which present two key challenges: (1) input values cannot be specified, precluding the use of pick-freeze estimators, and (2) the number of inputs can exceed $10^4$, preventing all input-output samples from being held simultaneously in memory on typical computational resources, thereby rendering conventional all-at-once implementations of given-data Sobol' index estimators impractical.
To address these challenges, we introduced a generalized given-data Sobol' index estimator that admits arbitrary partitions, a streaming algorithm that processes input-output samples in batches, an asymptotic analysis of the new estimator for zero-valued indices, and a practical screening heuristic motivated by this asymptotic behavior for identifying indices that cannot be distinguished from zero in finite samples.

These extensions were demonstrated on two analog neural network applications, where increased programming precision reduces network prediction variation but incurs an additional energy cost. 
Therefore, the goal in these problems is to identify a small subset of parameter to target for increased programming precision. 
The application exemplars were a satellite detection network with approximately $10^4$ input parameters, and a CIFAR10 classification network with approximately $2 \times 10^5$ input parameters.
In both applications, our approach identified a small subset of inputs with significant estimated first-order effects despite the extremely high input dimensionality.
This provides a practical basis for prioritizing programming-precision improvements subject to an energy budget.

Several directions remain for future work.
The generalized estimator and the partition error analyses in~\cref{sec:partition_accuracy} could inform adaptive partitioning methods that balance partition bias and statistical variance. 
Such methods may improve estimation of higher-order Sobol' indices. 
Existing given-data estimators for higher-order indices employ equiprobable partitions~\cite{zhai_space-partition_2014}, whereas the generalized estimator developed here permits arbitrary partitions. 
Because higher-order estimation requires partitioning the joint input space, it is substantially more sample-intensive, making adaptive partitioning strategies increasingly beneficial.
Extending the present framework to higher-order indices would be particularly valuable for analog neural networks, where interaction effects are expected to play an important role.

Several computational, empirical, and theoretical directions also warrant further investigation.
Although the update formulae in~\cref{eq:streaming_update_formulae} were introduced for one-pass streaming computation, they could equally be applied in distributed-memory settings; such implementations would facilitate iterative adaptive partition refinement, which generally requires multiple passes over the data and is therefore less suited to purely streaming computation.
Additional empirical evaluation  on a broader class of benchmark problems would strengthen understanding of partition schemes, partition initialization, and the screening heuristic; in particular, the Becker metafunction~\cite{puy_comprehensive_2022}, would provide a systematic means of evaluating the robustness of these design choices across a wide range of function and input behaviors.

Our asymptotic analyses motivate further investigation of estimator choice for applications in which many Sobol' indices are expected to be zero or near zero. 
While the implemented estimator performs well empirically for the proposed screening procedure, the leading-order effect of the unbiased variance estimators in the zero-index regime suggests that alternative estimator formulations may warrant consideration. 
Future work could systematically compare candidate estimators in terms of finite sample accuracy, zero-index behavior, and screening performance. 
Additionally, it would be interesting to investigate whether similar asymptotic behavior can be shown to exist for traditional pick-freeze estimators, motivating analogous theory-guided screening procedures.
\section*{Acknowledgements}
This work was supported by the Laboratory Directed Research and Development program
(Projects 233072 and 229363) at Sandia National Laboratories, a multimission laboratory managed and
operated by National Technology and Engineering Solutions of Sandia LLC, a wholly owned
subsidiary of Honeywell International Inc. for the U.S. Department of Energy’s National Nuclear
Security Administration under contract DE-NA0003525.
The authors further thank Drs. Pieterjan Robbe and Justin Winokur for their helpful suggestions on the paper.
This article has been authored by an employee of National Technology \& Engineering Solutions of Sandia, LLC. The employee owns all right, title and interest in and to the article and is solely responsible for its contents. The United States Government retains and the publisher, by accepting the article for publication, acknowledges that the United States Government retains a non-exclusive, paid-up, irrevocable, world-wide license to publish or reproduce the published form of this article or allow others to do so, for United States Government purposes. The DOE will provide public access to these results of federally sponsored research in accordance with the DOE Public Access Plan \url{https://www.energy.gov/downloads/doe-public-access-plan}.
This paper describes objective technical results and analysis. Any subjective views or opinions that might be expressed in the paper do not necessarily represent the views of the U.S. Department of Energy or the United States Government.

\appendix
\section{Asymptotic behavior of binned Sobol' indices}\label[appendix]{app:theory}
This appendix presents theoretical results on the asymptotic behavior of the binned Sobol' indices studied in this work.
Asymptotic consistency of the binned estimator was established in~\cite{borgonovo_common_2016} for a partition where the number of bins increases with sample size such that $\lim_{N\rightarrow\infty} \frac{N}{M(N)} = \infty$.
However, the streaming algorithm presented herein determines the partition during initialization and holds it fixed as samples are processed.
We therefore study the asymptotic behavior of the estimator as $N\rightarrow\infty$ for a fixed partition of size $M$.
Furthermore, we study the asymptotic behavior of the estimator for a zero-valued Sobol' index, which provides intuition for the sensitivity screening heuristic introduced in~\Cref{sec:heuristic_approach}.

Our analysis begins by establishing the consistency and asymptotic normality of a plug-in estimator expressed as a smooth function of empirical moments. 
We then study the special case of a zero-valued binned Sobol' index, showing that the first-order variation vanishes and deriving the corresponding second-order asymptotic distribution. 
Finally, we relate these results to the implemented estimator by proving that it differs from the plug-in estimator by a remainder of stochastic order $O_p(N^{-1})$.

\subsection{Preliminaries}
Throughout this appendix, we consider the fixed-partition Sobol' index
\begin{align}
    S_i^{(M)} := 1 - \frac{EV_i^{(M)}}{\V(f(\Xb))},
    \label{eq:partition_sobol}
\end{align}
where 
\begin{align}
    EV_i^{(M)} := \sum_{k=1}^M P(X_i \in A_k )\V( f(\Xb) | X_i \in A_k).
    \label{eq:partition_numerator}
\end{align}
Note that $S_i^{(M)}$ is only equal to $S_i$ as $M\rightarrow \infty$.

To facilitate the asymptotic analysis, define the random vector
    \begin{align*}
        \Zb := (I_1, I_1 f, I_1 f^2, \ldots, I_M, I_M f, I_M f^2, f, f^2),
    \end{align*}
where $I_k = \mathbbm{1}(X_i \in A_k)$ is the indicator random variable for bin $A_k$. 
Let its expectation be denoted
\begin{align*}
    \mu = \E[\Zb] = (p_1, m_{11}, m_{12}, \ldots, p_M, m_{M1}, m_{M2}, m_1, m_2), 
\end{align*}
and let the vector of empirical averages of $\Zb$ for sample size $N$ be denoted
\begin{align*}
    \EZ = (P_1, \hat{m}_{11}, \hat{m}_{12}, \ldots, P_M, \hat{m}_{M1}, \hat{m}_{M2}, \hat{m}_1, \hat{m}_2).
\end{align*}

Throughout this appendix, summations over bins are understood to be restricted to
\[
J=\{k:P(X_i\in A_k)>0\},
\]
since bins outside $J$ do not contribute to either $EV_i^{(M)}$ or any forms of its statistical estimators.

We define the smooth function $g$ so that
\begin{align}
    S_i^{(M)} &= g(\mu) = 1 - \frac{A(\mu)}{B(\mu)}, \label{eq:g_def}
\end{align}
where
\begin{align}
A(\mu) &= \sum_{k\in J}\left( m_{2k} - \frac{m_{1k}^2}{p_k} \right), \quad\quad
B(\mu) = m_2-m_1^2. \label{eq:g_A_B_def}
\end{align}
The corresponding plug-in estimator is defined as
\begin{align*}
    \tilde{S}_i^{(M)} = g(\EZ) = 1 - \frac{A(\EZ)}{B(\EZ)}.
\end{align*}
While this estimator is convenient for asymptotic analysis, it differs from the estimator defined in~\Cref{eq:general_estimator}, denoted $\hat{S}_i^{(M)}$ herein, which computes within-bin and total variances using unbiased sample variance estimators. 
Therefore, the asymptotic results below are first established for $ \tilde{S}_i^{(M)}$, and we then relate these results to $\hat{S}_i^{(M)}$ by characterizing the difference between the two estimators.

\subsection{Asymptotic properties of the plug-in estimator}

\begin{lemma}[Consistency]\label{lemma:consistency}
Assume that $\E[(f(\Xb))^2]<\infty$ and $\V(f(\Xb))>0$. Then
\begin{align*}
    \tilde{S}_i^{(M)}\xrightarrow{p} S_i^{(M)}.
\end{align*}
\end{lemma}
\begin{proof}
Since $\E[(f(\Xb))^2]<\infty$, every component of $\Zb$ has finite expectation. 
Then by the weak Law of Large Numbers $\EZ \xrightarrow{p} \mu$. 
Since $\V(f(\Xb))>0$, the denominator of $g$ is nonzero at $\mu$, and $g$ is continuous in a neighborhood of $\mu$. 
Therefore, the Continuous Mapping Theorem yields
\begin{align*}
    \tilde{S}_i^{(M)} = g(\hat{\Zb}) \xrightarrow{p} g(\mu) = S_i^{(M)}.
\end{align*}
\end{proof}

\begin{theorem}[Asymptotic normality]\label{thm:asymptotic_normality}
Assume that $\E[f(\Xb)^4]<\infty$ and $\V(f(\Xb))>0$. 
Then
\begin{align*}
    \sqrt{N}\left(\tilde S_i^{(M)}-S_i^{(M)}\right)
    \xrightarrow{d}
    \mathcal N(0,\sigma^2),
\end{align*}
where
\begin{align*} 
    \sigma^2 = \grad g(\mu)^\top \Sigma g(\mu),
\end{align*}
and $\Sigma=\Cov(\Zb)$.
\end{theorem}

\begin{proof}
Since $\E[f(\Xb)^4]<\infty$, every component of $\Zb$ has finite variance. 
Therefore, the multivariate Central Limit Theorem yields
\begin{align*}
    \sqrt{N}\left(\EZ-\mu\right)
    \xrightarrow{d} \Wb,
\end{align*}
where $\Wb= \mathcal N(0,\Sigma)$ with $\Sigma=\Cov(\Zb)$.
Since $\V(f(\Xb))>0$, the mapping $g$ is continuously differentiable in a neighborhood of $\mu$, and the result follows immediately from the multivariate delta method.
\end{proof}

\begin{remark}
The covariance matrix $\Sigma$ is generally singular because the components of $\Zb$ satisfy linear identities such as $\sum_{k=1}^M I_k = 1$. 
This does not affect the above result since the multivariate Central Limit Theorem and delta method remain valid when the limiting normal distribution is degenerate. 
\end{remark}

\begin{lemma}\label{lem:null_index}
    If the Sobol' index $S_i$ is zero, the corresponding fixed-partition Sobol' index $S_i^{(M)}$ is also zero.
\end{lemma}
\begin{proof}
        Since $S_i=0$
    \begin{align*}
        \E[f(\Xb)|X_i] = \E[f(\Xb)] \;{\rm a.s.}
    \end{align*} 
    By the law of total variance in each bin,
    \begin{align*}
        \V(f(\Xb)|X_i \in A_k ) = \E[\V(f(\Xb)|X_i) \,|\, X_i \in A_k ] + \V(\E[f(\Xb)|X_i] \,|\, X_i \in A_k),
    \end{align*}
    and the second term vanishes. 
    Then
    \begin{align*}
        EV_i^{(M)} &= \sum_{k=1}^M P(X_i \in A_k )\V(f(\Xb) | X_i \in A_k)  \\
        &= \sum_{k=1}^M P(X_i \in A_k ) \E[\V(f(\Xb) | X_i ) | X_i \in A_k] \\
        &=  \E[\V(f(\Xb)|X_i)].
    \end{align*}
    Since 
    \begin{align*}
        S_i = 1-\frac{\E[\V(f(\Xb) | X_i )]}{\V(f(\Xb))},
    \end{align*}
    the fact that $S_i=0$ implies 
     $ \E[\V(f(\Xb)|X_i)] = \V(f(\Xb))$. 
     Therefore, $EV_i^{(M)} = \V(f(\Xb))$, and consequently $S_i^{(M)} = 1 - \frac{\V(f(\Xb))}{\V(f(\Xb))} = 0$.
\end{proof}

\begin{lemma}[Vanishing first variation under zero index assumption]\label{lem:first_variation_null}
Suppose $S_i^{(M)}=0$. 
Then for every feasible perturbation of $\mu$
satisfying
\begin{align*}
\sum_k\delta p_k=0,\qquad
\sum_k\delta m_{k1}=\delta m_1,\qquad
\sum_k\delta m_{k2}=\delta m_2,
\end{align*}
we have
\begin{align*}
\delta g := D g(\mu)[\delta \mu] = 0.
\end{align*}
\end{lemma}

\begin{proof}

The first variation of $A$ is
\begin{align*}
    \delta A = \sum_{k} \left( 
        \delta m_{k2} - 2\frac{m_{k1}}{p_k}\delta m_{k1} + \frac{m^2_{k1}}{p_k^2} \delta p_k
    \right).
\end{align*}
Since $S_i^{(M)}=0$, we have that  $ \V(f(\Xb))=\sum_k p_k \V(f(\Xb)|X_i \in A_k)$.
By the law of total variance over the partition, this is only possible if $E[f(\Xb) | X_i \in A_k ] = \E[f(\Xb)]$ for every occupied bin, which is equivalent to $m_{k1} = p_k m_1$.
Then 
\begin{align*}
    \delta A = \sum_k \left( 
     \delta m_{k2} - 2 m_1 \delta m_{k1} + m_1^2 \delta p_k 
    \right).
\end{align*}
Applying the feasibility constraints yields
\begin{align*}
    \delta A = \delta m_2 - 2 m_1 \delta m_1.
\end{align*}
Differentiating $B$ gives
\begin{align*}
    \delta B = \delta m_2 - 2m_1 \delta m_1.
\end{align*}
Hence $\delta A = \delta B.$
Since $S_i^{(M)}=0$ we also have that $A=B$.
Therefore,
\begin{align*}
    \delta g = - \frac{\delta A}{B} + \frac{A}{B^2} \delta B =  \frac{\delta B - \delta A}{B} = 0.
\end{align*}
\end{proof}

\begin{theorem}[Zero-index asymptotics]\label{thm:zero_index_asymptotics}
    Suppose the assumptions of~\Cref{thm:asymptotic_normality} hold, and $S_i = 0$. 
    Then
    \begin{align*}
        N(\tilde{S}_i^{(M)} - S_i^{(M)}) = N \tilde{S}_i^{(M)} \xrightarrow{d} \frac{1}{2}\Wb^\top H(\mu)\Wb,
    \end{align*}
    where $\Wb\sim\mathcal{N}(0,\Sigma)$, $\Sigma=\Cov(\Zb)$, and $H(\cdot)$ is the Hessian of $g$.
    Notably,
    \begin{align*}
        \tilde{S}_i^{(M)} = O_p(N^{-1}).
    \end{align*}
\end{theorem}
\begin{proof}
By Lemma~\ref{lem:null_index},
\begin{align*}
    S_i^{(M)} = 0.
\end{align*}
Since $g$ is twice continuously differentiable in a neighborhood of $\mu$, Taylor's theorem gives the second-order expansion
\begin{align*}
    \tilde{S}_i^{(M)} = g(\EZ) = g(\mu) + Dg( \mu )[\delta\mu] + \frac{1}{2} \delta\mu^\top H(\tilde{\mu}) \delta\mu,
\end{align*}
where
\begin{align*}
    \delta\mu = \EZ - \mu
\end{align*}
and $\tilde{\mu} = \mu + h \delta \mu$ for some $h \in (0,1)$. 

Since the empirical moments satisfy
\begin{align*}
    \sum_k P_k = 1, \qquad
    \sum_k \widehat{m}_{k1} = \hat m_1, \qquad
    \sum_k \widehat{m}_{k2} = \hat m_2
\end{align*}
identically, the perturbation $\delta\mu$ satisfies the feasibility constraints of~\Cref{lem:first_variation_null}, and
\begin{align*}
    Dg(\mu)[\delta\mu] = 0.
\end{align*}
Since $S_i^{(M)} = g(\mu) = 0$, it follows that 
\begin{align*}
     \tilde S_i^{(M)} = \frac{1}{2}  (\,\delta\mu)^\top
    H(\tilde{\mu})
    (\,\delta\mu).
\end{align*}
Furthermore, since $\delta \mu = O_p(N^{-1/2})$, 
\begin{align*}
    \tilde{S}_i^{(M)} = O_p(N^{-1}).
\end{align*}
Multiplying by $N$ yields
\begin{align*}
    N\tilde S_i^{(M)} = \frac{1}{2}  (\sqrt N\,\delta\mu)^\top
    H(\tilde{\mu})
    (\sqrt N\,\delta\mu).
\end{align*}
By the multivariate Central Limit Theorem,
   $ \sqrt N\,\delta\mu \xrightarrow{d} \Wb,$
where $\Wb\sim\mathcal{N}(0,\Sigma)$. 
Since $\EZ \xrightarrow{p}\mu$, we also have $\tilde{\mu}\xrightarrow{p}\mu$. 
By continuity of the Hessian,
    $H(\tilde{\mu}) \xrightarrow{p} H(\mu).$
Applying Slutsky's theorem and the Continuous Mapping Theorem, we have
\begin{align*}
    N(\tilde S_i^{(M)} - S_i^{(M)}) = N\tilde S_i^{(M)} \xrightarrow{d} \frac12 \Wb^\top H(\mu)\Wb.
\end{align*}
\end{proof}

\subsection{Extension to the implemented estimator}

We now derive the difference from the plug-in estimator induced by the use of unbiased variance estimators in the implemented estimator. 
We first define the unbiased estimators in terms of the empirical moment vector $\EZ$. 
For the unbiased variance $\hat{V}$, 
\begin{align*}
    \hat{V} &= \frac{N}{N-1}(\hat m_2 - \hat m_1) = \frac{N}{N-1} B(\EZ). 
\end{align*}
Similarly for the within-bin variance $s_k^2$,
\begin{align*}
    s_k^2 = \frac{n_k}{n_k-1} \left( \frac{\hat m_{2k}}{P_k} - \frac{\hat m_{1k}^2}{P_k^2} \right) 
    = \frac{P_k N}{P_k N-1} \left( \frac{\hat m_{2k}}{P_k} - \frac{\hat m_{1k}^2}{P_k^2} \right).
\end{align*}

\begin{lemma}[Difference between plug-in and implemented estimator]\label{lem:implemented_diff}
    The implemented estimator with unbiased in-bin and total variance estimators is related to the plug-in estimator as
\begin{align*}
    \hat{S}_i^{(M)} = \tilde{S}_i^{(M)} + R_N,
\end{align*}
where 
\begin{align*}
    R_N = (\hat{m}_2 - \hat{m}_1^2)^{-1} \sum_k \frac{1-P_k}{N P_k - 1} \left( \hat{m}_{k2} - \frac{\hat{m}_{k1}^2}{P_k} \right).
\end{align*}
\end{lemma}

\begin{proof}
To simplify notation we denote the expressions in~\eqref{eq:g_A_B_def} for the plug-in estimator as $A$ and $B$ respectively.
The implemented estimator employs unbiased sample variance estimators within each bin and for the total variance. 
Hence
\begin{align*}
A_u &:= \sum_k P_k s_k^2 
= \sum_k P_k \frac{P_k N}{P_k N-1} \left( \frac{\hat m_{2k}}{P_k} - \frac{\hat m_{1k}^2}{P_k^2} \right) 
= \sum_k \frac{NP_k}{NP_k-1} \left( \hat m_{k2} - \frac{\hat m_{k1}^2}{P_k} \right),
\end{align*}
and
\begin{align*}
B_u := \hat V = \frac{N}{N-1}B.
\end{align*}
Therefore,
\begin{align*}
\hat S_i^{(M)} = 1-\frac{A_u}{B_u} = 1- \frac{N-1}{N} \frac{A_u}{B}.
\end{align*}
Now, since
\begin{align*}
\left(\frac{N-1}{N}\right)\left( \frac{NP_k}{NP_k-1} \right) = 1-\frac{1-P_k}{NP_k-1},
\end{align*}
we have that
\begin{align*}
\frac{N-1}{N}A_u &= \sum_k \left( 1- \frac{1-P_k}{NP_k-1} \right) \left( \hat m_{k2} - \frac{\hat m_{k1}^2}{P_k} \right) \\
&= A- \sum_k \frac{1-P_k}{NP_k-1} \left( \hat m_{k2} - \frac{\hat m_{k1}^2}{P_k} \right).
\end{align*}
Substituting this identity into the expression for
\(\hat S_i^{(M)}\) gives
\begin{align*}
\hat S_i^{(M)} &= 1-\frac{A}{B} + \frac{1}{B} \sum_k \frac{1-P_k}{NP_k-1} \left( \hat m_{k2} - \frac{\hat m_{k1}^2}{P_k} \right) \\
&= \tilde{S}_i^{(M)} + \frac{1}{B} \sum_k \frac{1-P_k}{NP_k-1} \left( \hat m_{k2} - \frac{\hat m_{k1}^2}{P_k} \right),
\end{align*}
yielding 
\begin{align*}
    R_N =  (\hat{m}_2 - \hat{m}_1^2 )^{-1} \sum_k \frac{1-P_k}{NP_k-1} \left( \hat m_{k2} - \frac{\hat m_{k1}^2}{P_k} \right).
\end{align*}
\end{proof}

\begin{proposition}[Asymptotic behavior of the difference]\label{prop:difference_asymptotics}
\begin{align*}
    N R_N \xrightarrow{p} c,
\end{align*}
where
\begin{align*}
    c = \frac{\sum_k (1 - p_k)\V(f(\Xb)|X_i \in A_k)}{\V(f(\Xb))}.
\end{align*}
\end{proposition}
\begin{proof}
By definition, 
\begin{align*}
N R_N &= \frac{1}{\hat m_2-\hat m_1^2} \sum_k \frac{N(1-P_k)}{NP_k-1} \left( \hat m_{k2} - \frac{\hat m_{k1}^2}{P_k} \right).
\end{align*}
Since $ P_k\xrightarrow{p} p_k > 0 $, we have
\begin{align*}
    \frac{N(1-P_k)}{NP_k-1} = \frac{1-P_k}{P_k-\frac{1}{N}}  \xrightarrow{p} \frac{1-p_k}{p_k}.
\end{align*}
Additionally, by the Weak Law of Large Numbers and the Continuous Mapping Theorem, we have
\begin{align*}
   \hat m_{k2}-\frac{\hat m_{k1}^2}{P_k} \xrightarrow{p} p_k\V(f(\Xb)| X_i \in A_k)
\end{align*}
and
\begin{align*}
    \hat m_2-\hat m_1^2 \xrightarrow{p} \V(f(\Xb)).
\end{align*}
Since the number of bins is fixed, by Slutsky's theorem
\begin{align*}
N R_N &\xrightarrow{p} \frac{1}{\V(f(\Xb))} \sum_k \frac{1-p_k}{p_k} (p_k\V(f(\Xb)| X_i \in A_k)) \\
&= \frac{1}{\V(f(\Xb))} \sum_k (1-p_k)\V(f(\Xb)| X_i \in A_k).
\end{align*}
\end{proof}

\begin{remark}
For nonzero indices, $R_N$ is asymptotically negligible relative to the $O_p(N^{-1/2})$ estimation error of the implemented and plug-in estimators.
Thus, although the use of unbiased sample variance estimators introduces a correction of order $O_p(N^{-1})$, it does not affect the first-order asymptotic distribution for $S_i>0$.
\end{remark}

\begin{theorem}[Zero-index asymptotics for the implemented estimator]\label{thm:implemented_zero_asymptotics}
Suppose the assumptions of \Cref{thm:asymptotic_normality} hold, and that $S_i=0$.
Then
\begin{align*}
    N\hat S_i^{(M)}
    \xrightarrow{d}
    \frac12 \Wb^\top H(\mu)\Wb + c,
\end{align*}
where $\Wb\sim\mathcal N(0,\Sigma)$, $H(\mu)$ is the Hessian of $g$ evaluated at $\mu$, and
\begin{align*}
    c = \frac{1}{\V(f(\Xb))} \sum_k (1-p_k) \V(f(\Xb) | X_i \in A_k).
\end{align*}
In particular,
\begin{align*}
    \hat S_i^{(M)}=O_p(N^{-1}).
\end{align*}
\end{theorem}
\begin{proof}
By \Cref{thm:zero_index_asymptotics},
\begin{align*}
N\tilde S_i^{(M)}
\xrightarrow{d}
\frac12\Wb^\top H(\mu)\Wb,
\end{align*}
while \Cref{prop:difference_asymptotics} gives
\begin{align*}
NR_N\xrightarrow{p}c.
\end{align*}
Since
\begin{align*}
\hat S_i^{(M)} = \tilde S_i^{(M)} + R_N,
\end{align*}
Slutsky's theorem implies
\begin{align*}
N\hat S_i^{(M)}
&=
N\tilde S_i^{(M)}
+
NR_N \xrightarrow{d}
\frac12\Wb^\top H(\mu)\Wb+c.
\end{align*}
Finally,
\[
N\hat S_i^{(M)}=O_p(1),
\]
which is equivalent to
\[
\hat S_i^{(M)}=O_p(N^{-1}).
\]
\end{proof}

\begin{remark}\label{rem:positive_shift}
For a nontrivial partition (i.e., at least one occupied bin has positive conditional variance), the constant $c$ is positive, since $1-p_k \geq 0$, $\V(f(\Xb) | X_i \in A_k) \geq 0$, and $V(f(\Xb)) > 0$.
For a zero-valued index, the $O_p(N^{-1})$ remainder $R_N$ is on the same scale as the leading-order fluctuation of the plug-in estimator and therefore contributes a positive shift $c$ to the limiting distribution of the implemented estimator.
For an equiprobable partition, $p_k=1/M$, and the variance decomposition
$V(f(\Xb))=\sum_k p_k \V(f(\Xb)|X_i \in A_k)$ gives $c=M-1$.
\end{remark}

\begin{remark}\label[remark]{rem:skewness}
    The limiting distribution in~\Cref{thm:zero_index_asymptotics} is a quadratic form in a multivariate Gaussian random vector, which admits the representation 
    \begin{align*}
    \frac12 \Wb^\top H(\mu)\Wb \overset{d} = \frac12 \sum_{j=1}^r \lambda_j \chi_{1,j}^2,
    \end{align*}
    where $\lambda_j$ are the nonzero eigenvalues of $\Sigma^{1/2} H(\mu)\Sigma^{1/2}$, $r = \mathrm{rank}(\Sigma^{1/2} H(\mu)\Sigma^{1/2})$, and $\chi_{1,j}^2$ are independent chi-squared random variables with one degree of freedom. 
    Thus, the asymptotic distribution of the zero-index estimator is generally non-Gaussian and may exhibit positive skewness.
    However, since the moment vector satisfies three linear identities, the covariance matrix $\Sigma$ has rank at most $3M-1$, and therefore $r\leq 3M-1$. 
    Consequently, finer partitions permit a larger number of independent chi-squared contributions to the limiting distribution. 
    Therefore, although weighted sums of chi-squared random variables are generally positively skewed, one may expect the asymptotic skewness to become less pronounced as the number of bins increases, with a significant number of potential terms ($\sim$150-300) resulting from bin counts in the range of 50--100 considered in this work.

    By~\Cref{thm:implemented_zero_asymptotics}, the limiting distribution of the implemented estimator differs from this quadratic-form distribution only by the additive constant $c$. 
    As discussed in~\Cref{rem:positive_shift}, the use of the unbiased sample variance estimators shifts the limiting distribution to the right without changing its shape.
    Importantly, both the quadratic-form and shift contributions to the limiting distribution occur on the $N^{-1}$ scale.
    Thus, under the zero-index assumption, the estimator concentrates toward zero as $N$ increases while retaining a non-Gaussian, generally right-skewed null distribution.
\end{remark}
\section{Test problem full definitions}\label[appendix]{app:test_problems}
This appendix provides the detailed definition of the test functions introduced in~\cref{sec:test_problems}.

\subsection{Sobol' G function}
We employ the definition of the Sobol' G function from~\cite{saltelli_variance_2010}, defined for uniform random variable inputs 
$X_i\sim\mathcal{U}[0,1]$ as:
\begin{equation}
\begin{aligned}
    f_G(\Xb) &= \prod_{i=1}^{d} g_i, \quad g_i = \frac{\abs{4X_i - 2}+a_i}{1+a_i}, \quad a_i = \sqrt{i-1}, \\
    S_i  \V(f_G) = V_i &= \frac{1/3}{(1+a_i)^2}, \quad \V(f_G) = \prod_{i=1}^d (1+V_i) - 1.
\end{aligned}
    \label{eq:sobol_g}
\end{equation}
For test cases with 3 parameters, we set $a_1 = 0$, $a_2=0.5$, and $a_3=10000$. 
For test cases with 1000 parameters, we set $a_i = \sqrt{i-1}, i=1,\ldots,1000$, such that only a small subset of variables have significant Sobol' indices even for large numbers of inputs.

\subsection{Ishigami function}
We employ the definition of the Ishigami function from~\cite{marrel_calculations_2009}.
For uniform random variables $X_i\sim\mathcal{U}[-\pi,\pi], \, i=1,2,3$, the function and its Sobol indices are defined as:
\begin{equation*}
\begin{aligned}
    f_I(\Xb) &= \sin(X_1) + a \sin^2(X_2) + b X_3^4\sin(X_1), \\
    \V(f_I) &= \frac{a^2}{8} + \frac{b(\pi^4)}{5} + \frac{b^2\pi^8}{18} + 0.5 \\
    S_1\V(f_I) &= 0.5(1+b\pi^{4/5})^2, \quad\quad
    S_2\V(f_I) = a^2 / 8, \quad\quad
    S_3\V(f_I) = 0.
\end{aligned}
\end{equation*}

\subsection{Polynomial function}
The polynomial function is defined as
\begin{align*}
    f_P(\Xb) = aX_1 + bX_2^2 + cX_1 X_2 + 0 X_3,
\end{align*}
where $a,b,c$ are defined according to the random variable type.
The function is defined for uniform, normal, exponential, and spike-slab random variables.
The coefficients are defined as $[a,b,c]=[1,1,10]$ for uniform and exponential random variables and as $[a,b,c]=[1,1,1]$ for normal random variables.
The analytical expressions for Sobol' indices for these three distributions are reported in~\cref{tab:polynomial_sobols}.
$S_3=0$ for all distribution types.
\begin{table}[h!]
    \renewcommand{\arraystretch}{1.5}
    \centering
    \begin{tabular}{c|c|c|c}
        \toprule
        R.V. Distribution & $\V(f_P)$ & $S_1 \V(f_P)$ & $S_2\V(f_P)$ \\
        \hline
        $\mathcal{U}[0,1]$ & 
            $\frac{a^2}{12} + \frac{ac}{12} + \frac{4b^2}{45} + \frac{bc}{12} + \frac{7c^2}{144}$ &
            $\frac{a^2}{12} + \frac{ac}{12} + \frac{c^2}{48}$ &
            $\frac{4b^2}{45} + \frac{bc}{12} + \frac{c^2}{48}$ \\
       $\mathcal{N}(0,1)$ & 
            $a^2 + 2b^2 + c^2$ & 
            $a^2$ & 
            $2b^2$ \\
       $\exp(\lambda=1)$ & 
        $a^2 + 2ac + 20b^2 + 8bc + 3c^2$& 
        $a^2 + 2ac + c^2$ & 
        $20b^2 + 8bc + c^2$ \\
        \bottomrule
    \end{tabular}
    \caption{A table of the analytical values of Sobol' indices for the polynomial test function.}
    \label{tab:polynomial_sobols}
\end{table}

For the spike-slab distribution, we selected $p=0.5$, $[\mu,\sigma]=[1,1]$, and $[a,b,c]=[1,1,1]$.
It is more complex to analytically compute the Sobol' indices for the spike-slab random variable, so we computed numerical values using $10^7$ samples, which was enough to ensure at least two significant digits of accuracy.
These numerical indices, to two significant digits, are: $S_1=0.21$, $S_2=0.72$, and $S_3=0.0$.
\section{Analog neural network architectures}\label[apendix]{app:nn_architectures}
This appendix provides more in-depth descriptions of the analog networks studied in~\cref{sec:application}.

\subsection{Satellite detection network}
\begin{figure}[h]
    \centering
    \includegraphics[scale=0.7]{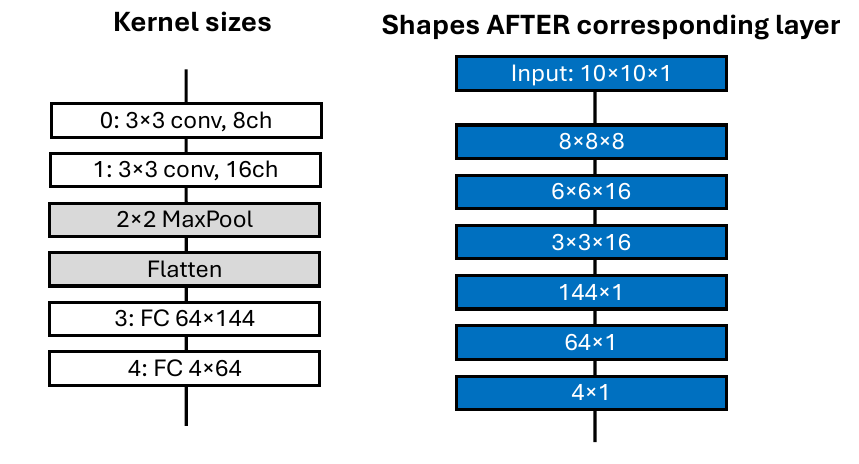} \vspace{-1em}
    \caption{Satellite detection network architecture. Left: kernel sizes in each layer. Right: shapes of the data after going through each layer.}
    \label{fig:satnetarch} 
\end{figure}

\label[appendix]{app:satnet_architecture}
Structurally, the satellite network has four layers: two convolutional and two densely connected layers as indicated in \cref{fig:satnetarch}.

To interpret this figure, remember that the input images are $10 \times 10$ pixel grayscale images, which means there is only one input channel. 
Layer 0 applies a $3 \times 3$ convolutional stencil to these images, so that there are 9 inputs to layer 0.
Technically, there is also a bias term as an additional input. 
However, we ignore the bias terms here in the count of GSA inputs because in this particular application these terms are implemented digitally so they are not subject to the inaccuracies seen in the weights that are represented with analog conductances.
Layer 0 has 8 output channels.
As this convolutional stencil sweeps over the full input image without zero-padding, layer 0 transforms the input images from $10 \times 10 \times 1$ to $8 \times 8 \times 8$.
The output of layer 0 gets passed to layer 1 for another $3 \times 3$ convolutional stencil application.
The input size for a convolutional stencil in layer 1 is therefore $3 \times 3 \times 8 = 72$ as there are 8 channels coming from layer 0.
Layer 1 has 16 output channels, so sweeping the convolutional stencil over the $8 \times 8 \times 8$ output from layer 0 (no zero-padding) gives an output of $6 \times 6 \times 16$.
After applying a $2 \times 2$ maxpool operation, this is reduced to $3 \times 3 \times 16$, which gets unrolled into a 144 element vector that becomes the input to layer 2.
Layer 2 is a fully connected layer that maps the 144 inputs into 64 outputs.
Layer 3 similarly maps its 64 inputs into the 4 outputs of the network.
Accordingly, \cref{tab:sat_arch} lists the number of weights in each layer of the network, adding up to a total of 10696.

\begin{table}[h]
\centering
\caption{Architecture of the satellite network with the number of weights in each layer. Bias terms are not included in this count as they are implemented without programming error.\label{tab:sat_arch}}
\begin{tabular}{|c|c|}
\hline
\textbf{Layer} & \textbf{Size} \\ 
& (inputs $\times$ outputs) \\
\hline
0 & 9 $\times$ 8 \\
\hline
1 & 72 $\times$ 16 \\
\hline
2 & 144 $\times$ 64 \\
\hline
3 & 64 $\times$ 4 \\
\hline\hline
Total & 10696 \\
\hline
\end{tabular}
\end{table}

\subsection{CIFAR-10 Image Classification Network}\label[appendix]{sec:cifar_architecture}

\begin{figure}[h]
    \centering
    \includegraphics[scale=0.7]{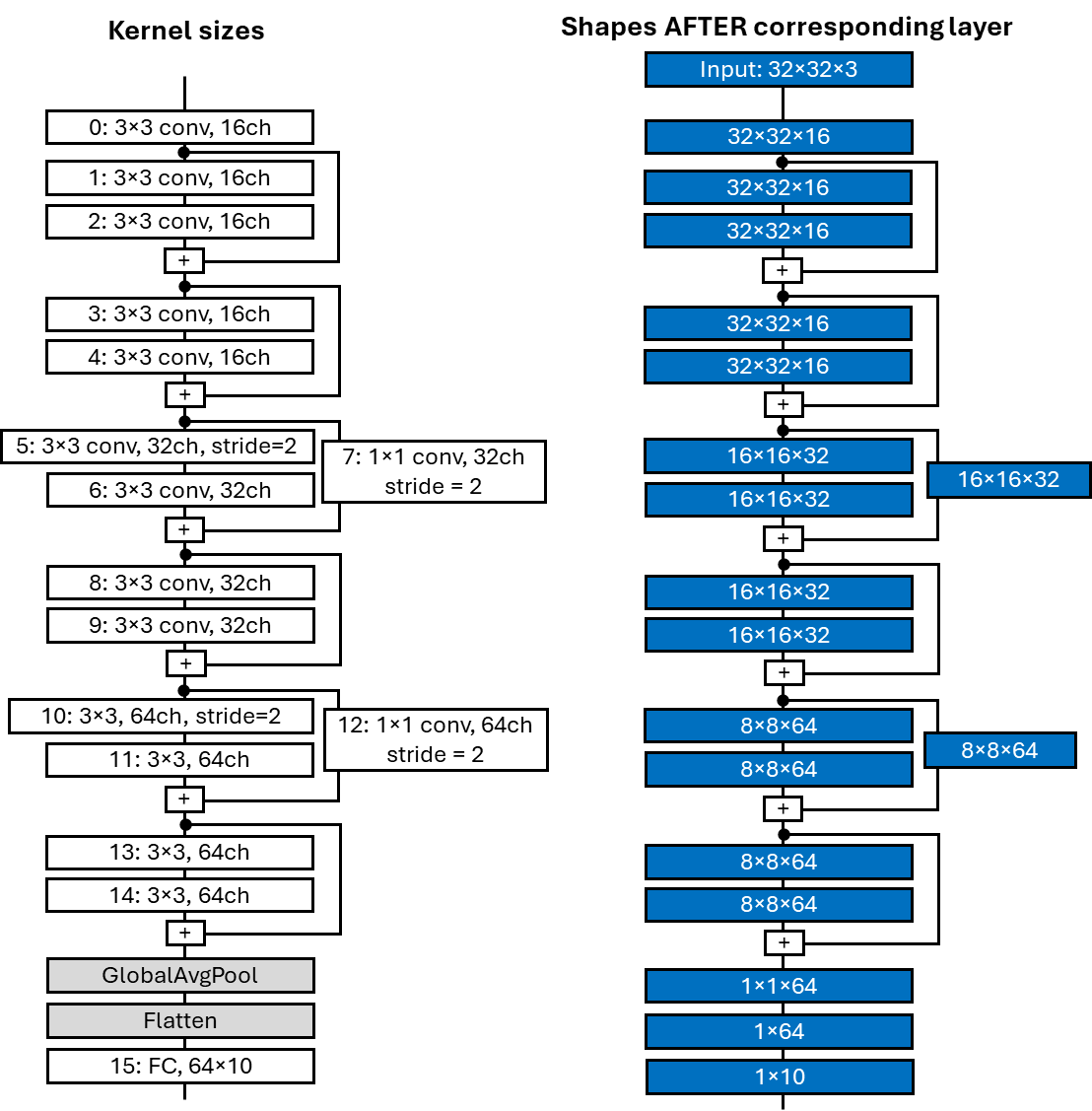} \vspace{-1em}
    \caption{ResNet-14 network architecture used to classify the CIFAR-10 images. Left: kernel sizes and strides in each layer. Right: shapes of the data after going through each layer.}
    \label{fig:resnetarch} 
\end{figure}

As shown in \cref{fig:resnetarch}, this network has 16 layers, 13 of which are $3 \times 3$ convolutional layers, two of them (layers 7 and 12) are projection shortcuts, and one (layer 15) is a fully connected layer.

Contrary to the satellite network in the previous section, the convolutions in this network use zero padding.
As such, the shape of the data is not altered as it passes through each layer, as illustrated on the right hand side of \cref{fig:resnetarch}, starting from the shape of the input images that are $32 \times 32 \times 3$.
However, layers 5 and 10 (and the associated shortcut layers 7 and 12) operate with stride 2, so that they halve the size in the pixel dimensions.
Near the end, a Global Average Pooling layer averages the pixel dimensions, and the result gets flattened into a 64 channel vector, which then gets mapped with a fully connected layer to a vector of 10 numbers. 
These numbers give the probability of the input image belonging to the corresponding CIFAR-10 class.

Following the same reasoning as with the satellite network, we can compute the number of weights in each layer, as listed in \cref{tab:cifar10_arch} with a total of 174128 weights.

\begin{table}[h]
\centering
\caption{Number of weights in each layer of the ResNet model. Bias terms are not included in this count as they are implemented without programming error. \label{tab:cifar10_arch}}
\begin{tabular}{|c|c|}
\hline
\textbf{Layer} & \textbf{Size} \\ 
& (inputs $\times$ outputs) \\
\hline
0         & 27 $\times$ 16 \\
\hline
1 --- 4   & 144 $\times$ 16 \\
\hline
5         & 144 $\times$ 32 \\
\hline
6         & 288 $\times$ 32 \\
\hline
7         & 16 $\times$ 32 \\
\hline
8 --- 9   & 288 $\times$ 32 \\
\hline
10        & 288 $\times$ 64 \\
\hline
11        & 576 $\times$ 64 \\
\hline
12        & 32 $\times$ 64 \\
\hline
13 --- 14 & 576 $\times$ 64 \\
\hline
15        & 64 $\times$ 10 \\
\hline\hline
Total & 174128 \\
\hline
\end{tabular}
\end{table}

\clearpage

\singlespacing
\nocite{*} 
\bibliographystyle{unsrturl} 
\bibliography{references}

@article{razavi_future_2021,
	title = {The {Future} of {Sensitivity} {Analysis}: {An} essential discipline for systems modeling and policy support},
	volume = {137},
	issn = {1364-8152},
	url = {https://www.sciencedirect.com/science/article/pii/S1364815220310112},
	doi = {10.1016/j.envsoft.2020.104954},
	journal = {Environmental Modelling \& Software},
	author = {Razavi, Saman and Jakeman, Anthony and Saltelli, Andrea and Prieur, Clémentine and Iooss, Bertrand and Borgonovo, Emanuele and Plischke, Elmar and Lo Piano, Samuele and Iwanaga, Takuya and Becker, William and Tarantola, Stefano and Guillaume, Joseph H.A. and Jakeman, John and Gupta, Hoshin and Melillo, Nicola and Rabitti, Giovanni and Chabridon, Vincent and Duan, Qingyun and Sun, Xifu and Smith, Stefán and Sheikholeslami, Razi and Hosseini, Nasim and Asadzadeh, Masoud and Puy, Arnald and Kucherenko, Sergei and Maier, Holger R.},
	month = mar,
	year = {2021},
	pages = {104954},
}

@incollection{iooss_review_2015,
	address = {Boston, MA},
	title = {A {Review} on {Global} {Sensitivity} {Analysis} {Methods}},
	isbn = {978-1-4899-7547-8},
	url = {https://doi.org/10.1007/978-1-4899-7547-8_5},
	booktitle = {Uncertainty {Management} in {Simulation}-{Optimization} of {Complex} {Systems}: {Algorithms} and {Applications}},
	publisher = {Springer US},
	author = {Iooss, Bertrand and Lemaître, Paul},
	editor = {Dellino, Gabriella and Meloni, Carlo},
	year = {2015},
	doi = {10.1007/978-1-4899-7547-8_5},
	pages = {101--122},
}

@article{helton_survey_2006,
	title = {Survey of sampling-based methods for uncertainty and sensitivity analysis},
	volume = {91},
	issn = {0951-8320},
	url = {https://www.sciencedirect.com/science/article/pii/S0951832005002292},
	doi = {10.1016/j.ress.2005.11.017},
	number = {10},
	journal = {The Fourth International Conference on Sensitivity Analysis of Model Output (SAMO 2004)},
	author = {Helton, J.C. and Johnson, J.D. and Sallaberry, C.J. and Storlie, C.B.},
	month = oct,
	year = {2006},
	pages = {1175--1209},
}

@article{borgonovo_common_2016,
	title = {A {Common} {Rationale} for {Global} {Sensitivity} {Measures} and {Their} {Estimation}},
	volume = {36},
	issn = {0272-4332},
	url = {https://doi.org/10.1111/risa.12555},
	doi = {10.1111/risa.12555},
	number = {10},
	urldate = {2024-11-13},
	journal = {Risk Analysis},
	author = {Borgonovo, Emanuele and Hazen, Gordon B. and Plischke, Elmar},
	month = oct,
	year = {2016},
	note = {Publisher: John Wiley \& Sons, Ltd},
	pages = {1871--1895},
}

@article{li_efficient_2016,
	title = {An efficient modularized sample-based method to estimate the first-order {Sobol}' index},
	volume = {153},
	issn = {0951-8320},
	url = {https://www.sciencedirect.com/science/article/pii/S0951832016300266},
	doi = {10.1016/j.ress.2016.04.012},
	journal = {Reliability Engineering \& System Safety},
	author = {Li, Chenzhao and Mahadevan, Sankaran},
	month = sep,
	year = {2016},
	pages = {110--121},
}

@article{plischke_global_2013,
	title = {Global sensitivity measures from given data},
	volume = {226},
	issn = {0377-2217},
	url = {https://www.sciencedirect.com/science/article/pii/S0377221712008995},
	doi = {10.1016/j.ejor.2012.11.047},
	number = {3},
	journal = {European Journal of Operational Research},
	author = {Plischke, Elmar and Borgonovo, Emanuele and Smith, Curtis L.},
	month = may,
	year = {2013},
	pages = {536--550}
}

@article{zhai_space-partition_2014,
	title = {Space-partition method for the variance-based sensitivity analysis: {Optimal} partition scheme and comparative study},
	volume = {131},
	issn = {0951-8320},
	url = {https://www.sciencedirect.com/science/article/pii/S0951832014001434},
	doi = {10.1016/j.ress.2014.06.013},
	journal = {Reliability Engineering \& System Safety},
	author = {Zhai, Qingqing and Yang, Jun and Zhao, Yu},
	month = nov,
	year = {2014},
	pages = {66--82}
}

@article{saltelli_making_2002,
	title = {Making best use of model evaluations to compute sensitivity indices},
	volume = {145},
	issn = {0010-4655},
	url = {https://www.sciencedirect.com/science/article/pii/S0010465502002801},
	doi = {10.1016/S0010-4655(02)00280-1},
	number = {2},
	journal = {Computer Physics Communications},
	author = {Saltelli, Andrea},
	month = may,
	year = {2002},
	pages = {280--297},
}

@book{saltelli_global_2007,
	title = {Global {Sensitivity} {Analysis}. {The} {Primer}},
	isbn = {978-0-470-72518-4},
	url = {https://onlinelibrary.wiley.com/doi/book/10.1002/9780470725184},
	publisher = {John Wiley \& Sons, Ltd},
	author = {Saltelli, Andrea and Ratto, Marco and Andres, Terry and Campolongo, Francesca and Cariboni, Jessica and Gatelli, Debora and Saisana, Michaela and Tarantola, Stefano},
	month = dec,
	year = {2007},
}

@article{saltelli_variance_2010,
	title = {Variance based sensitivity analysis of model output. {Design} and estimator for the total sensitivity index},
	volume = {181},
	issn = {0010-4655},
	url = {https://www.sciencedirect.com/science/article/pii/S0010465509003087},
	doi = {10.1016/j.cpc.2009.09.018},
	number = {2},
	journal = {Computer Physics Communications},
	author = {Saltelli, Andrea and Annoni, Paola and Azzini, Ivano and Campolongo, Francesca and Ratto, Marco and Tarantola, Stefano},
	month = feb,
	year = {2010},
	pages = {259--270},
}

@article{marrel_calculations_2009,
	title = {Calculations of {Sobol} indices for the {Gaussian} process metamodel},
	volume = {94},
	issn = {0951-8320},
	url = {https://www.sciencedirect.com/science/article/pii/S0951832008001981},
	doi = {10.1016/j.ress.2008.07.008},
	number = {3},
	journal = {Reliability Engineering \& System Safety},
	author = {Marrel, Amandine and Iooss, Bertrand and Laurent, Béatrice and Roustant, Olivier},
	month = mar,
	year = {2009},
	pages = {742--751},
}

@incollection{prieur_variance-based_2017,
	address = {Cham},
	title = {Variance-{Based} {Sensitivity} {Analysis}: {Theory} and {Estimation} {Algorithms}},
	isbn = {978-3-319-12385-1},
	url = {https://link.springer.com/content/pdf/10.1007%2F978-3-319-12385-1_35.pdf?pdf=core},
	booktitle = {Handbook of {Uncertainty} {Quantification}},
	publisher = {Springer International Publishing},
	author = {Prieur, Clémentine and Tarantola, Stefano},
	editor = {Ghanem, Roger and Higdon, David and Owhadi, Houman},
	year = {2017},
	doi = {10.1007/978-3-319-12385-1_35},
	pages = {1217--1239},
}

@article{sobol_sensitivity_1993,
	title = {Sensitivity estimates for nonlinear mathematical models},
	volume = {1},
	url = {http://www.andreasaltelli.eu/file/repository/sobol1993.pdf},
	number = {4},
	journal = {Math. Model. Comput. Exp},
	author = {Sobol, I. M.},
	year = {1993},
	pages = {407--414},
}

@article{sobol_global_2001,
	title = {Global sensitivity indices for nonlinear mathematical models and their {Monte} {Carlo} estimates},
	volume = {55},
	issn = {0378-4754},
	url = {https://www.sciencedirect.com/science/article/pii/S0378475400002706},
	doi = {10.1016/S0378-4754(00)00270-6},
	number = {1},
	journal = {The Second IMACS Seminar on Monte Carlo Methods},
	author = {Sobol, I.M},
	month = feb,
	year = {2001},
	pages = {271--280},
}

@inproceedings{chan_updating_1982,
	address = {Heidelberg},
	title = {Updating {Formulae} and a {Pairwise} {Algorithm} for {Computing} {Sample} {Variances}},
	isbn = {978-3-642-51461-6},
	url = {https://doi.org/10.1007/978-3-642-51461-6_3},
	doi = {10.1007/978-3-642-51461-6_3},
	booktitle = {{COMPSTAT} 1982 5th {Symposium} held at {Toulouse} 1982},
	publisher = {Physica-Verlag HD},
	author = {Chan, T. F. and Golub, G. H. and LeVeque, R. J.},
	editor = {Caussinus, H. and Ettinger, P. and Tomassone, R.},
	year = {1982},
	pages = {30--41},
}

@techreport{pebay_formulas_2008,
	address = {United States},
	title = {Formulas for robust, one-pass parallel computation of covariances and arbitrary-order statistical moments.},
	url = {https://www.osti.gov/biblio/1028931},
	language = {English},
	number = {SAND2008-6212},
	author = {Pebay, Philippe Pierre},
	year = {2008},
	doi = {10.2172/1028931},
}

@article{hyndman_sample_1996,
	title = {Sample {Quantiles} in {Statistical} {Packages}},
	volume = {50},
	issn = {00031305},
	url = {http://www.jstor.org/stable/2684934},
	doi = {10.2307/2684934},
	number = {4},
	urldate = {2025-04-30},
	journal = {The American Statistician},
	author = {Hyndman, Rob J. and Fan, Yanan},
	year = {1996},
	note = {Publisher: [American Statistical Association, Taylor \& Francis, Ltd.]},
	pages = {361--365},
}

@misc{CrossSim,
  author={Ben Feinberg and T. Patrick Xiao and Curtis J. Brinker and Christopher H. Bennett and Matthew J. Marinella and Sapan Agarwal},
  title={{CrossSim: accuracy simulation of analog in-memory computing}},
  url = {https://github.com/sandialabs/cross-sim},
}

@misc{CrossSimModels,
  title={{CrossSim Pre-Trained Models}},
  url = {https://github.com/sandialabs/cross-sim-models},
}

@inproceedings{Krizhevsky2009,
  title={Learning Multiple Layers of Features from Tiny Images},
  author={Alex Krizhevsky},
  year={2009},
  url={https://api.semanticscholar.org/CorpusID:18268744}
}

@article{Xiao:2020,
    author = {Xiao, T. Patrick and Bennett, Christopher H. and Feinberg, Ben and Agarwal, Sapan and Marinella, Matthew J.},
    title = {Analog architectures for neural network acceleration based on non-volatile memory},
    journal = {Applied Physics Reviews},
    volume = {7},
    number = {3},
    pages = {031301},
    year = {2020},
    month = {07},
    issn = {1931-9401},
    doi = {10.1063/1.5143815},
    url = {https://doi.org/10.1063/1.5143815},
}

@ARTICLE{Xiao:2022,
  author={Xiao, T. Patrick and Feinberg, Ben and Bennett, Christopher H. and Prabhakar, Venkatraman and Saxena, Prashant and Agrawal, Vineet and Agarwal, Sapan and Marinella, Matthew J.},
  journal={IEEE Circuits and Systems Magazine}, 
  title={On the Accuracy of Analog Neural Network Inference Accelerators}, 
  year={2022},
  volume={22},
  number={4},
  pages={26-48},
  doi={10.1109/MCAS.2022.3214409}
}

@INPROCEEDINGS{Agrawal:2022,
  author={Agrawal, V. and Xiao, T. P. and Bennett, C. H. and Feinberg, B. and Shetty, S. and Ramkumar, K. and Medu, H. and Thekkekara, K. and Chettuvetty, R. and Leshner, S. and Luzada, Z. and Hinh, L. and Phan, T. and Marinella, M. J. and Agarwal, S.},
  booktitle={2022 International Electron Devices Meeting (IEDM)}, 
  title={Subthreshold operation of SONOS analog memory to enable accurate low-power neural network inference}, 
  year={2022},
  volume={},
  number={},
  pages={21.7.1-21.7.4},
  doi={10.1109/IEDM45625.2022.10019564}
}

@INPROCEEDINGS{Xiao:2023,
  author={Xiao, T. P. and Wahby, W. S. and Bennett, C. H. and Hays, P. and Agrawal, V. and Marinella, M. J. and Agarwal, S.},
  booktitle={2023 IEEE Symposium on VLSI Technology and Circuits (VLSI Technology and Circuits)}, 
  title={Enabling High-Speed, High-Resolution Space-based Focal Plane Arrays with Analog In-Memory Computing}, 
  year={2023},
  volume={},
  number={},
  pages={1-2},
  doi={10.23919/VLSITechnologyandCir57934.2023.10185348}
}

@INPROCEEDINGS{He:2016,
  author={He, Kaiming and Zhang, Xiangyu and Ren, Shaoqing and Sun, Jian},
  booktitle={2016 IEEE Conference on Computer Vision and Pattern Recognition (CVPR)}, 
  title={Deep Residual Learning for Image Recognition}, 
  year={2016},
  volume={},
  number={},
  pages={770-778},
  doi={10.1109/CVPR.2016.90}
}

@article{lo_piano_uncertainty_2022,
	title = {Uncertainty appraisal provides useful information for the management of a manual grape harvest},
	volume = {219},
	issn = {1537-5110},
	url = {https://www.sciencedirect.com/science/article/pii/S1537511022001167},
	doi = {10.1016/j.biosystemseng.2022.05.006},
	journal = {Biosystems Engineering},
	author = {Lo Piano, Samuele and Parenti, Alessandro and Guerrini, Lorenzo},
	month = jul,
	year = {2022},
	pages = {259--267},
}

@article{campolongo_effective_2007,
	title = {An effective screening design for sensitivity analysis of large models},
	volume = {22},
	issn = {1364-8152},
	url = {https://www.sciencedirect.com/science/article/pii/S1364815206002805},
	doi = {10.1016/j.envsoft.2006.10.004},
	number = {10},
	journal = {Modelling, computer-assisted simulations, and mapping of dangerous phenomena for hazard assessment},
	author = {Campolongo, Francesca and Cariboni, Jessica and Saltelli, Andrea},
	month = oct,
	year = {2007},
	pages = {1509--1518},
}

@article{b_ciuffo_sensitivity-analysis-based_2014,
	title = {A {Sensitivity}-{Analysis}-{Based} {Approach} for the {Calibration} of {Traffic} {Simulation} {Models}},
	volume = {15},
	issn = {1558-0016},
	url = {https://doi.org/10.1109/TITS.2014.2302674},
	doi = {10.1109/TITS.2014.2302674},
	number = {3},
	journal = {IEEE Transactions on Intelligent Transportation Systems},
	author = {{B. Ciuffo} and {C. Lima Azevedo}},
	month = jun,
	year = {2014},
	pages = {1298--1309},
}

@article{sheikholeslami_global_2019,
	title = {Global sensitivity analysis for high-dimensional problems: {How} to objectively group factors and measure robustness and convergence while reducing computational cost},
	volume = {111},
	issn = {1364-8152},
	url = {https://www.sciencedirect.com/science/article/pii/S1364815218301464},
	doi = {10.1016/j.envsoft.2018.09.002},
	journal = {Environmental Modelling \& Software},
	author = {Sheikholeslami, Razi and Razavi, Saman and Gupta, Hoshin V. and Becker, William and Haghnegahdar, Amin},
	month = jan,
	year = {2019},
	pages = {282--299},
}

@article{lo_piano_variance-based_2021,
	title = {Variance-based sensitivity analysis: {The} quest for better estimators and designs between explorativity and economy},
	volume = {206},
	issn = {0951-8320},
	url = {https://www.sciencedirect.com/science/article/pii/S0951832020307961},
	doi = {10.1016/j.ress.2020.107300},
	journal = {Reliability Engineering \& System Safety},
	author = {Lo Piano, Samuele and Ferretti, Federico and Puy, Arnald and Albrecht, Daniel and Saltelli, Andrea},
	month = feb,
	year = {2021},
	pages = {107300},
}

@article{archer_sensitivity_1997,
	title = {Sensitivity measures, {ANOVA}-like {Techniques} and the use of bootstrap},
	volume = {58},
	issn = {0094-9655},
	url = {https://doi.org/10.1080/00949659708811825},
	doi = {10.1080/00949659708811825},
	number = {2},
	journal = {Journal of Statistical Computation and Simulation},
	publisher = {Taylor \& Francis},
	author = {Archer, G. E. B. and Saltelli, A. and Sobol, I. M.},
	month = may,
	year = {1997},
	pages = {99--120},
}

@article{gamboa_statistical_2016,
	title = {Statistical inference for {Sobol} pick-freeze {Monte} {Carlo} method},
	volume = {50},
	issn = {0233-1888},
	url = {https://doi.org/10.1080/02331888.2015.1105803},
	doi = {10.1080/02331888.2015.1105803},
	number = {4},
	journal = {Statistics},
	publisher = {Taylor \& Francis},
	author = {Gamboa, F. and Janon, A. and Klein, T. and Lagnoux, A. and Prieur, C.},
	month = jul,
	year = {2016},
	pages = {881--902},
}

@article{azzini_sobol_2021,
	title = {Sobol’ main effect index: an {Innovative} {Algorithm} ({IA}) using {Dynamic} {Adaptive} {Variances}},
	volume = {213},
	issn = {0951-8320},
	url = {https://www.sciencedirect.com/science/article/pii/S0951832021001885},
	doi = {10.1016/j.ress.2021.107647},
	journal = {Reliability Engineering \& System Safety},
	author = {Azzini, Ivano and Rosati, Rossana},
	month = sep,
	year = {2021},
	pages = {107647},
}

@article{puy_comprehensive_2022,
	title = {A {Comprehensive} {Comparison} of {Total}-{Order} {Estimators} for {Global} {Sensitivity} {Analysis}},
	volume = {12},
	url = {https://www.dl.begellhouse.com/journals/52034eb04b657aea,221f21d635d3d68f,6de6348a17770dba.html},
	doi = {10.1615/Int.J.UncertaintyQuantification.2021038133},
	number = {2},
	publisher = {Begell House},
	author = {Puy, Arnald and Becker, William and Piano, Samuele Lo and Saltelli, Andrea},
	year = {2022},
	note = {Type: 10.1615/Int.J.UncertaintyQuantification.2021038133},
	pages = {1--18},
}
\end{document}